\documentclass{article}
\PassOptionsToPackage{numbers, compress}{natbib}
\usepackage[main,final]{neurips_2026}

\usepackage[utf8]{inputenc}
\usepackage[T1]{fontenc}
\usepackage{hyperref}
\usepackage{url}
\usepackage{booktabs}
\usepackage{amsfonts}
\usepackage{amsmath}
\usepackage{amssymb}
\usepackage{nicefrac}
\usepackage{microtype}
\usepackage{xcolor}
\usepackage{graphicx}
\usepackage{enumitem}
\usepackage{placeins}
\usepackage{multirow}
\usepackage{tabularx}
\usepackage{algorithm}
\usepackage{algpseudocode}
\algrenewcommand\algorithmicrequire{\textbf{Input:}}
\algrenewcommand\algorithmicensure{\textbf{Output:}}

\newcommand{\method}{\textsc{CoSE}}
\newcommand{\methodlong}{Compositional Sparse Embedding}
\newcommand{\sysname}{\textsc{CHARM}}

\let\normalpm\pm
\newcommand{\smalluncertainties}{%
  \renewcommand{\pm}{\scriptstyle\normalpm}%
}

\title{Structured Sparse Memory for Recurrent Reasoning}

\author{%
  Zixuan Zhao \\
  Department of Computer Science\\
  University of Chicago\\
  Chicago, IL 60637 \\
  \texttt{vapor@uchicago.edu} \\
  \And
  Samuel Wheeler \\
  Argonne National Laboratory \\
  Lemont, IL 60439 \\
  \texttt{swheeler@anl.gov} \\
  \And
  Neil Getty \\
  Argonne National Laboratory \\
  Lemont, IL 60439 \\
  \texttt{ngetty@anl.gov} \\
  \AND
  Xiaotian Duan \\
  Argonne National Laboratory \\
  Lemont, IL 60439 \\
  \texttt{duan@anl.gov} \\
  \And
  Rick Stevens \\
  Argonne National Laboratory \\
  Lemont, IL 60439 \\
  \texttt{stevens@anl.gov} \\
  \And
  Fangfang Xia \\
  Argonne National Laboratory \\
  Lemont, IL 60439 \\
  \texttt{fangfang@anl.gov} \\
}

\begin{document}

{\setlength{\tabcolsep}{4pt}\maketitle}

\begin{abstract}

Recurrent models trained from scratch have recently become competitive on ARC-style reasoning tasks, but the usual framing around small recurrent backbones overlooks two important parts of the system: task-conditioned memory and synthetic augmentation data. We study this regime through CHARM, a compact hybrid ARC model that combines recurrent reasoning with structured task memory, synthetic data, and inference-time aggregation. In existing approaches, task-conditioned memory supplies a large hidden source of capacity, reaching more than 30x the size of the recurrent backbone. We introduce a compositional sparse embedding (\method{}) for task conditioning that reduces learned task-memory parameters by over 90\% while improving pass@2 in controlled ARC ablations. We also introduce Re-ARC2, a set of programs that generate and verify new examples for all ARC-AGI-2 training tasks. Combining these ingredients, our system reaches 84\% pass@2 on ARC-AGI-1 and 46.7\% pass@2 on ARC-AGI-2 public evaluation. Structured memory also enables adaptation with only 512 parameters per unseen puzzle, and its benefits generalize to other domains. Our code, dataset, and model checkpoints are available at \url{https://github.com/water-vapor/charm}.
\end{abstract}

\section{Introduction}
\label{sec:intro}

Recurrent models have emerged as an alternative paradigm for neural reasoning. Unlike chain-of-thought methods, which express intermediate steps as autoregressive text, these systems perform latent-space reasoning through repeated computation, iteratively refining hidden states or candidate solutions. These models are small enough to train end-to-end within the task domain, so any reasoning capability they exhibit does not depend on language or vision priors inherited from pretraining. Empirically, this approach has been especially effective for problems requiring iterative hypothesis updating or structured transformation, including Sudoku, maze solving, and the ARC-AGI challenge~\citep{wang2025hrm,jolicoeur2025trm,gao2025urm}.

We study this regime in ARC-AGI, where a recent line of from-scratch systems has achieved strong results. This includes the recurrent HRM, TRM, and URM family, which we refer to as XRM, as well as the vision-based VARC system~\citep{wang2025hrm,jolicoeur2025trm,gao2025urm,hu2025varc}. These systems are often discussed through a common contrast with frontier-model ARC solutions, with emphasis on their compact backbone sizes of 7--27M parameters. But what actually drives performance remains unsettled: recent analyses suggest that performance can come from less visible ingredients such as outer-loop refinement and training-time task augmentation \cite{arcprize2025hrmanalysis, jolicoeur2025trm}. They also point to task identity embeddings as a critical but underanalyzed source of task conditioning \cite{arcprize2025hrmanalysis}.

\begin{figure}[!t]
  \centering
  \includegraphics[width=\linewidth]{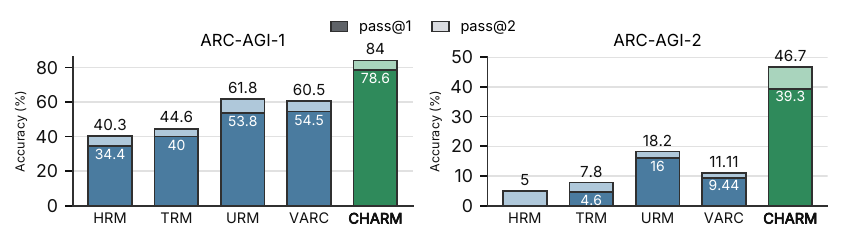}
  \caption{\textbf{Public-evaluation accuracy of ARC systems trained from scratch.} \sysname{} achieves the highest accuracy in this regime on both ARC-AGI-1 and ARC-AGI-2. The reported numbers are full-system results, combining structured task memory, synthetic data, and inference-time aggregation.}
  \label{fig:overall-arc-performance}
\end{figure}

In existing XRM systems, the task embedding table can be more than 30 times larger than the backbone itself: $\sim$448M learnable parameters on ARC-AGI-1 and $\sim$610M on ARC-AGI-2. Yet it is often overlooked because it is sparse and excluded from model-size counts. Removing or aggressively shrinking it, however, causes large accuracy drops.

This motivates a closer look at what the table stores and what structure it should have. In the stock design, each puzzle-augmentation instance indexes an independent learned vector, which is supplied to the recurrent backbone on every forward pass. This gives the model substantial instance-specific capacity, but it ignores the known structure of the index: puzzle identity, dihedral transform, and color permutation. We therefore ask whether task memory can be made more compositional without losing the instance-specific capacity that makes the full table effective.

On performance, we also find that synthetic data can outweigh recurrent architectural choices. Synthetic data takes augmentation one step further: instead of rotating, flipping, or recoloring existing examples, it generates new input-output pairs from the underlying task rule. While XRM systems use only symmetry-based task augmentations~\citep{wang2025hrm,jolicoeur2025trm,gao2025urm}, VARC uses Re-ARC synthetic data~\citep{hu2025varc,hodel2024rearc}; we extend that idea to ARC-AGI-2 with matched Re-ARC2 generators. In our experiments, this matched synthetic data is the largest measured driver of ARC-AGI-2 gains.

We study these interactions through \sysname{}, an integrated ARC system that combines an XRM-style recurrent backbone with structured task memory and synthetic data. Its design balances memory capacity, augmentation structure, recurrent reasoning, and inference-time compute, giving us a common experimental framework for studying how these ingredients affect performance. Our contributions are:

\begin{itemize}[leftmargin=*,itemsep=0pt,topsep=2pt]
  \item We provide an empirical decomposition of from-scratch recurrent ARC performance, showing that accuracy depends on three main components: recurrent compute, task-conditioned memory, and rule-preserving synthetic data (Table~\ref{tab:component-grid}).
  \item We show that the sparse task embedding table, indexed by puzzle and augmentation, dominates the parameter count and functions as task-conditioned memory; naive factorization over puzzle identity, dihedral transform, and color permutation consistently leaves a capability gap.
  \item We introduce structured sparse memory, which closes this gap by combining compositional task structure with a low-rank instance residual. The resulting module, \method{} (\methodlong{}), uses $1/15$ of the learned task-memory parameters while improving accuracy.
  \item We curate Re-ARC2, a synthetic training set extending Re-ARC--style rule-preserving generation to ARC-AGI-2, covering all 1{,}000 training tasks. 
  \item We generalize the structured sparse memory design to cellular automata reasoning and language modeling with Engram~\citep{cheng2026engram} memory modules. It also enables continual puzzle adaptation with just 512 parameters per task and fixed shared weights.
  \item \sysname{} integrates these ingredients while balancing recurrent reasoning depth with the truncated learning horizon, reaching 84\% pass@2 on ARC-AGI-1 and 46.7\% pass@2 on ARC-AGI-2 public evaluation with extended training.\footnote{We extend the training for our best configuration to around 1.6M steps on ARC-AGI-1 and 2.4M steps on ARC-AGI-2 to near-convergence and report final accuracy. Other ARC ablations use budgets of 600k and 1M steps, respectively, and report best accuracy with final 10-checkpoint voting unless otherwise stated.}
\end{itemize}

\begin{table}[t]
  \centering
  \caption{\textbf{Major contributors to \sysname{} performance.} We report public-evaluation pass@2 accuracy and total learned parameters varying task memory and synthetic data under the default evaluator. Values are percentage means $\pm$ sample standard deviations over 3--4 runs. Training budgets are 600k steps for ARC-AGI-1 and 1M for ARC-AGI-2.}
  \label{tab:component-grid}
  \small
  \smalluncertainties
  \begin{tabular}{llrrrr}
    \toprule
    \multirow{2}{*}{Benchmark} & \multirow{2}{*}{Synthetic data} & \multicolumn{2}{c}{Full task table} & \multicolumn{2}{c}{\method{}}\\
    \cmidrule(lr){3-4}\cmidrule(lr){5-6}
     & & pass@2 & Params & pass@2 & Params\\
    \midrule
    \multirow{2}{*}{ARC-AGI-1} & N/A & $59.97 \pm 1.77$ & 462M & $64.91 \pm 2.25$ & 43M\\
                               & Re-ARC & $78.50 \pm 1.03$ & 463M & $\mathbf{80.71} \pm 0.51$ & 43M\\
    \midrule
    \multirow{2}{*}{ARC-AGI-2} & N/A & $15.42 \pm 1.51$ & 623M & $18.61 \pm 1.63$ & 53M\\
                               & Re-ARC2 & $28.23 \pm 3.58$ & 674M & $\mathbf{37.22} \pm 1.77$ & 57M\\
    \bottomrule
  \end{tabular}
\end{table}

\section{Background and related work}
\label{sec:background}

\textbf{ARC and specialized reasoning systems.}
The Abstraction and Reasoning Corpus (ARC) was introduced to test whether systems can infer new abstract rules from only a few examples~\citep{chollet2019measure}. Each task provides a small set of input-output demonstrations and asks the solver to produce the output for one or two held-out inputs governed by the same hidden transformation. ARC-AGI-1 contains 400 training, 400 public evaluation, and 100 private test tasks; ARC-AGI-2 expands the training set to 1{,}000 tasks and provides 120 public evaluation and 120 private test tasks~\citep{chollet2025arc2}. We report pass@$k$, where a task is counted correct if the true output appears among the top $k$ predictions.

Recent specialized ARC systems trained from scratch include the recurrent XRM line (HRM~\citep{wang2025hrm}, TRM~\citep{jolicoeur2025trm}, and URM~\citep{gao2025urm}), as well as the vision-based VARC system~\citep{hu2025varc}. XRM systems use compact recurrent compute blocks with deep supervision and task-conditioned embeddings; VARC treats ARC as image-to-image translation with test-time training and ensembling. We follow the XRM evaluation setup, where public-evaluation training pairs are available during training but held-out outputs are not.

\textbf{Adaptation, augmentation, and test-time compute.}
ARC systems often rely on extra computation or data beyond a single forward pass. LoRA-based and Kaggle-style systems such as ARChitects~\citep{architects2024} adapt models on the demonstration pairs of a test task. VARC similarly uses test-time training with multi-view ensembling. XRM-style systems instead keep weights fixed and aggregate predictions across augmented views, recurrent depths, and recent checkpoints; our evaluator follows this pattern, with an unweighted recent-checkpoint window.

Data expansion is a separate source of gains. Simple ARC augmentations rotate, reflect, or recolor existing examples while preserving the underlying rule. Re-ARC~\citep{hodel2024rearc} goes further by providing generators and verifiers for ARC-AGI-1 training tasks, making it possible to sample new examples from the same rules. ConceptARC~\citep{moskvichev2023conceptarc} provides additional ARC-style tasks organized by concept. We use these public datasets for ARC-AGI-1 and curate Re-ARC2, an analogous synthetic set covering ARC-AGI-2's 1{,}000 training tasks.

\textbf{Conditional memory and structured embeddings.}
Conditional memory separates what a model computes from what it retrieves. Recent language-model work such as Engram~\citep{cheng2026engram} studies scalable lookup memories keyed by token contexts. The XRM task table is a simpler but extreme instance of this pattern: a static lookup indexed by task and augmentation, with the table holding the bulk of the learned parameters. ARC makes this setting unusually structured because the key decomposes into puzzle identity, dihedral transform, and color permutation.

Our replacement draws on structured embedding and modulation ideas. FiLM-style conditioning~\citep{perez2018film} uses learned affine modulation to inject side information into a representation, while hash and factorized embeddings~\citep{hashembed} reduce large lookup tables by sharing parameters across structured keys. In contrast to generic compression settings, our goal is to preserve the instance-specific capacity of the XRM task table while exposing the known compositional structure of ARC augmentations.

\section{Structured task memory}
\label{sec:task-memory}

\begin{figure}[ht]
  \centering
  \begin{minipage}[t]{0.49\linewidth}\centering
    \includegraphics[width=\linewidth]{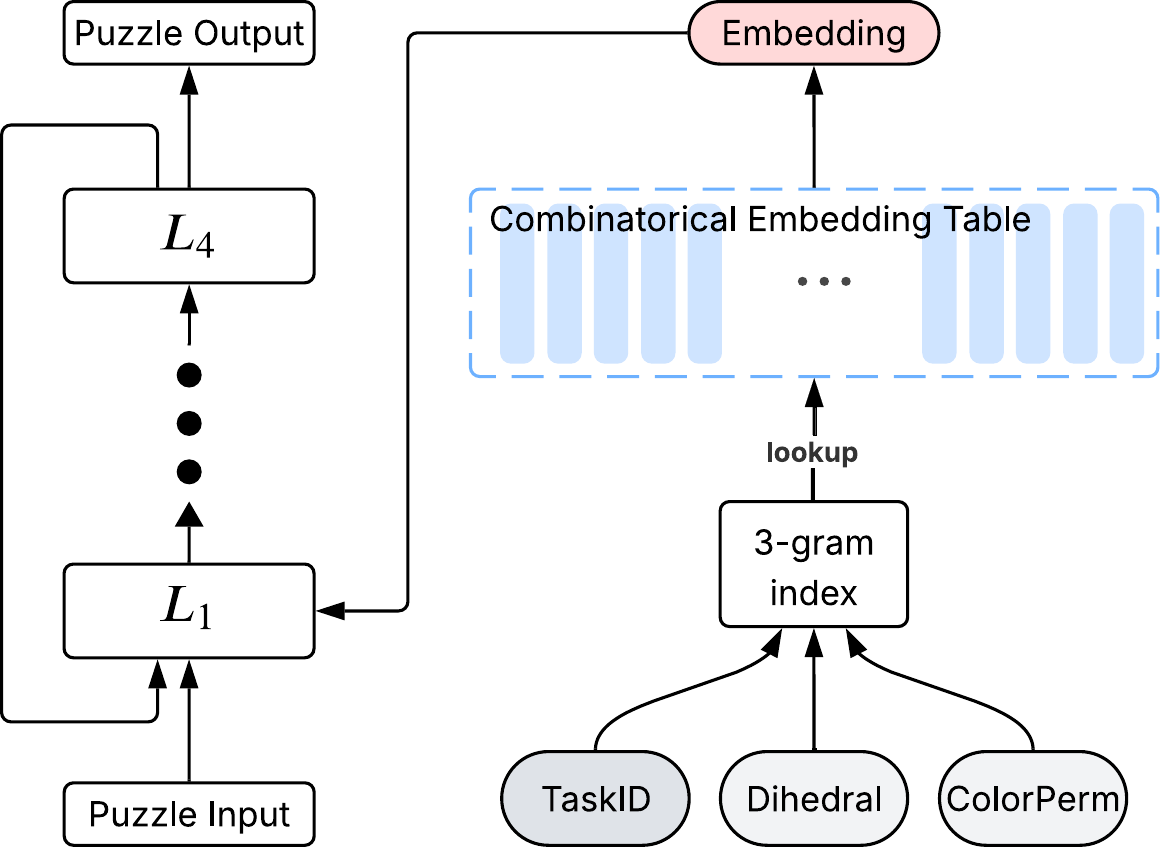}\\
    \footnotesize (a) original backbone and task table
  \end{minipage}\hfill
  \begin{minipage}[t]{0.49\linewidth}\centering
    \includegraphics[width=\linewidth]{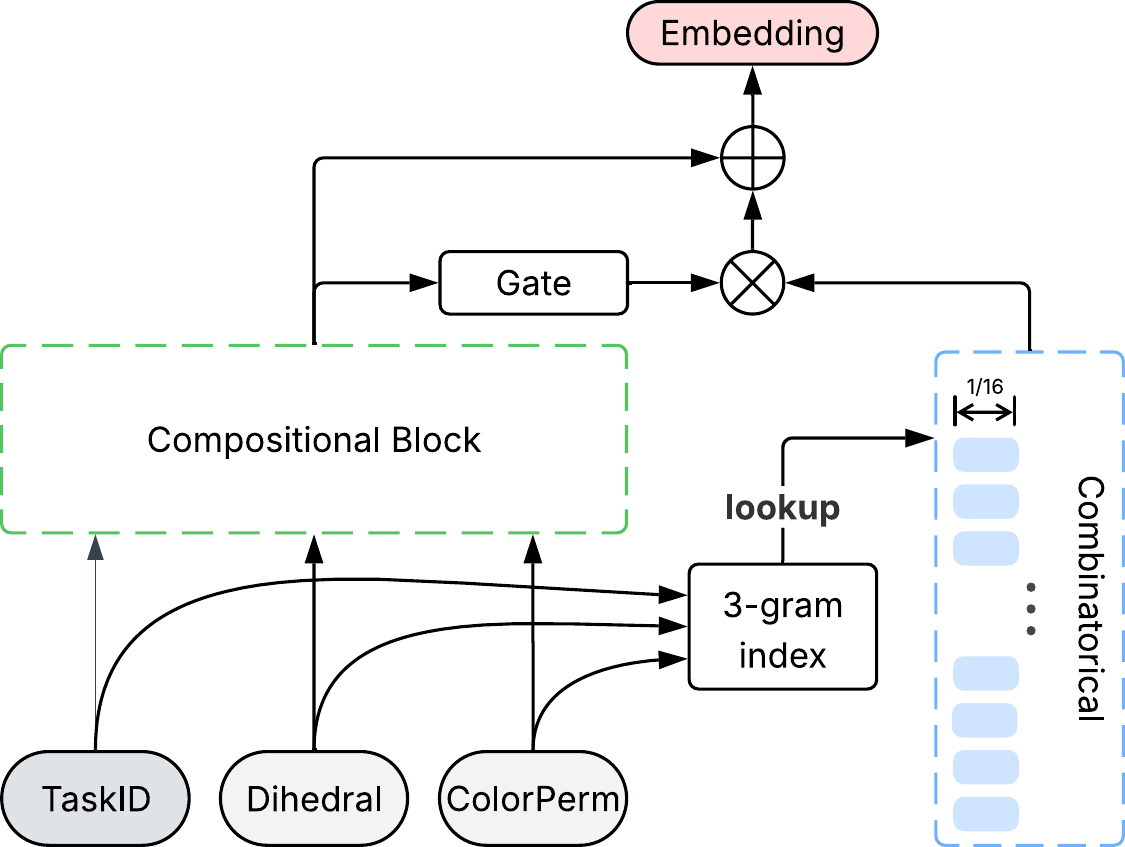}\\
    \footnotesize (b) our factorized task memory with residual
  \end{minipage}
  \caption{\textbf{From full task table to \methodlong{} (\method{}).}
  The stock lookup table assigns an independent full-width vector to each combination of puzzle identity, geometric transform, and color permutation.
  \method{} replaces it with two complementary branches: a compositional branch that factorizes these task properties and a per-instance residual branch with $1/16$ the original row width.
  A conditional gate merges the two branches into the task embedding passed to the recurrent backbone.
  }
  \label{fig:architecture-comparison}
\end{figure}

This section studies the task-memory component behind XRM systems. We begin by showing that the standard sparse task table is not a small implementation detail but the dominant learned parameter store. We then test two natural ways to shrink it and show why each fails on its own. These failures motivate our structured sparse memory design (Figure~\ref{fig:architecture-comparison}).

\subsection{Task tables dominate from-scratch ARC parameter counts}

The original ARC-AGI-1 dataset contains only around 1{,}000 puzzle pairs, which is insufficient to train models with a few million parameters. Existing XRM methods augment data using dihedral transformations and color permutations~\citep{wang2025hrm,jolicoeur2025trm,gao2025urm}, though they may not preserve the underlying rules. Concretely, every training example carries a descriptor $(i,g,\pi)$, where $i\in\{1,\ldots,P\}$ is the puzzle id, $g\in D_4$ is the dihedral augmentation, and $\pi\in S_9$ is the color permutation over the non-background colors. XRM works then convert this tuple into a single instance index $j=\mathrm{idx}(i,g,\pi)$ and retrieve a learned embedding row:
\begin{equation}
  e^\text{full}_j = E^\text{full}[j] \in \mathbb{R}^{d}, \qquad
  E^\text{full}\in\mathbb{R}^{N_\text{inst}\times d}.
  \label{eq:fulltable}
\end{equation}

The resulting table contains one row for each stored puzzle$\times$augmentation instance. In our measured ARC-AGI-1 settings, $N_\text{inst}\approx0.876$M and $d=512$, giving about $448$M task-table parameters. In ARC-AGI-2, $N_\text{inst}\approx1.19$M and the table has about $610$M parameters. This is an order of magnitude larger than the recurrent backbone.

From a representation learning perspective, the table is inherently redundant. The same puzzle under a rotated view or color permutation should not require a completely distinct and unrelated representation, since substantial structure is shared across related augmentations. Conversely, each per-instance embedding is specific to its corresponding instance and cannot be reused or shared across related examples. 

\subsection{Structure alone and compression alone are insufficient}

\paragraph{The structured compositional embedding design}

To improve parameter efficiency, we first factorize the task properties and compose them using a small set of MLP projectors. We learn separate representations for the puzzle id, the dihedral transform, and the color permutation:
\begin{align}
  p_i &= P[i]\in\mathbb{R}^{d}, &
  u_g &= G[g]\in\mathbb{R}^{d}, &
  c_\pi &= \mathrm{SlotPerm}(\pi;\, b,C_1,\ldots,C_9)\in\mathbb{R}^{d}.
  \label{eq:factors}
\end{align}
$P\in\mathbb{R}^{P\times d}$ is a puzzle embedding table and $G\in\mathbb{R}^{8\times d}$ is a dihedral embedding table. The color-permutation embedding stores a base vector $b$ and nine color-slot vectors $C_1,\ldots,C_9$, with $d_\text{base}+9d_\text{slot}=d$. For a permutation $\pi$, we read the slots in permuted order:
\begin{equation}
  c_\pi = \big[\,b\,\big|\,C_{\pi(1)}\,\big|\,\cdots\,\big|\,C_{\pi(9)}\,\big].
\end{equation}
This covers all $9!=362{,}880$ color permutations with one shared set of slot parameters.

The dihedral and color embeddings are concatenated and used to modulate the puzzle embedding with a FiLM-style operator~\citep{perez2018film}. Let $a_{g,\pi}=[u_g;c_\pi]$. Then
\begin{equation}
  \gamma_{g,\pi} = W_\gamma a_{g,\pi}+b_\gamma,\qquad
  \beta_{g,\pi}  = W_\beta  a_{g,\pi}+b_\beta,\qquad
  e^\text{comp}_{i,g,\pi} = (1+\gamma_{g,\pi})\odot p_i + \beta_{g,\pi}.
  \label{eq:cose-comp}
\end{equation}
We also experimented with several alternative composition operators, including addition, FiLM variants, hypernetworks, and mixture-of-experts modules. While some alternatives achieve performance close to FiLM, we observe a persistent gap relative to the original full embedding table (see Table \ref{tab:compo-ablations} in Appendix~\ref{app:composition-ablations}).

We focus on the pass@1000 metric to evaluate the capability of these methods, since it is less sensitive to training noise and better reflects whether a model can produce the correct solution at all. On the ARC-AGI-1 training split, we observe roughly a ~7\% gap between the compositional methods and the full embedding table. In particular, there exists a set of 19 puzzles that can be solved by the full embedding table, but by none of the purely compositional methods. Although we were unable to manually identify any shared pattern or puzzle type within this subset, the result suggests a fundamental capability gap rather than a mere difference in statistical success rate.

\paragraph{Composition is not enough: capacity matters}

We design an experiment that tests whether the pure-composition gap comes from the combination operator or from lost memory capacity. We compare two embedding designs under two data augmentation regimes. The full combinatorial table assigns an independent vector to each puzzle-augmentation instance. The separated-table baseline learns independent puzzle, dihedral, and color-permutation embeddings and adds them elementwise. Under ordinary unrestricted color permutations, the color table has up to $9!=362{,}880$ entries, giving the separated setup capacity comparable to the full table. Under the pooled regime, color permutations are restricted to a fixed set of 1{,}000, reducing the separated color table while leaving the underlying task distribution nearly unchanged for the full table. The performance drop we observed in Table \ref{tab:cppool} therefore isolates capacity as a major source of the factorization gap.

\begin{table}[ht]
  \centering
  \caption[Color-permutation pooling isolates the capacity effect in task embeddings.]{\textbf{Color-permutation pooling isolates the capacity effect in task embeddings.} A fixed pool of 1{,}000 color permutations leaves the performance of the full combinatorial table nearly unchanged but sharply hurts separated puzzle, dihedral, and color-permutation embeddings.\protect\footnotemark}
  \xdef\arcOneTrainingFootnote{\thefootnote}
  \label{tab:cppool}
  \small
  \smalluncertainties
  \begin{tabular}{p{0.36\linewidth}p{0.22\linewidth}cc}
    \toprule
    Embedding setup & Color permutation & pass@2 & pass@1000\\
    \midrule
    Combinatorial task table & Unrestricted & $49.75 \pm 2.50$ & $71.75 \pm 0.22$\\
    \midrule
    Combinatorial task table & 1000-pooled & 52.08 & 71.25\\
    3 separate tables & Unrestricted & 52.33 & 71.00\\
    3 separate tables & 1000-pooled & $45.52 \pm 0.80$ & $62.19 \pm 0.97$\\
    \bottomrule
  \end{tabular}
\end{table}
\footnotetext{\label{fn:arc1train}Models are trained on the ARC-AGI-1 training set for 170k steps. Reported standard deviations are specific to this short-training split setting. 600k-step public-evaluation runs are substantially more stable.}

\paragraph{Low-rank-only compression degrades performance}

We also explore a more direct approach to reduce parameter count by compressing the embedding-table bandwidth. Specifically, we reduce the width of each embedding vector from 512 to smaller dimensions and project it back to the model dimension after lookup. Table \ref{tab:lowrank-ablations} shows that accuracy degrades progressively as the embedding rank is reduced, suggesting that this approach alone does not provide an obvious path towards parameter reduction without sacrificing capability.

\begin{table}[ht]
  \centering
  \caption[Standalone low-rank task tables lose accuracy as row width shrinks.]{\textbf{Standalone low-rank task tables lose accuracy as row width shrinks.} The stock row uses the full-width TRM task table with $r=512$; the remaining rows replace it with a narrower per-instance table without the compositional branch.\protect\footnotemark[\arcOneTrainingFootnote]}
  \label{tab:lowrank-ablations}
  \small
  \smalluncertainties
  \begin{tabular}{p{0.54\linewidth}cc}
    \toprule
    Embedding setup & pass@2 & pass@1000\\
    \midrule
    Stock TRM full task table ($r=512$) & $49.75 \pm 2.50$ & $71.75 \pm 0.22$\\
    \midrule
    Low-rank task table ($r=128$) & 47.71 & 69.38\\
    Low-rank task table ($r=64$) & 46.04 & 66.00\\
    Low-rank task table ($r=32$) & 45.46 & 67.50\\
    Low-rank task table ($r=8$) & $38.29 \pm 0.95$ & $55.92 \pm 1.35$\\
    \bottomrule
  \end{tabular}
\end{table}

\subsection{\method{} combines compositional structure with residual memory}

Motivated by the observations above, we introduce \method{}, which combines the structured compositional embedding with a low-rank residual table. 

\paragraph{Architectural Design.} The compositional branch produces $e^\text{comp}_{i,g,\pi}$  as in Equation~\ref{eq:cose-comp}, the residual branch keeps a small table $R\in\mathbb{R}^{N_\text{inst}\times r}$ with $r=32$ by default. It is indexed by the same instance id $j=\mathrm{idx}(i,g,\pi)$ as the full table, then projected back to hidden width:
\begin{equation}
  \rho_j = U R[j],\qquad U\in\mathbb{R}^{d\times r}.
\end{equation}
The final task embedding is
\begin{equation}
  e^\text{\method{}}_{i,g,\pi}
  = e^\text{comp}_{i,g,\pi} + m_{i,g,\pi}\odot \rho_j,
  \qquad
  m_{i,g,\pi}=\sigma(W_m e^\text{comp}_{i,g,\pi}+b_m),
  \label{eq:cose-final}
\end{equation}
where the sigmoid gate is optional. Without the gate, $m\equiv\mathbf{1}$. In the default setting, the residual table has $1/16$ the row width of the stock task table. The full architecture with the compositional block can be found in Figure \ref{fig:cose-detail} in Appendix~\ref{app:cose-module}; alternative residual designs are compared in Appendix~\ref{app:cose-ablations}.

\paragraph{Results.} Our \method{} uses only roughly 1/15 the parameters of the original full embedding table across ARC-AGI setups (Appendix~\ref{app:param-count} reports the full parameter accounting). Despite this substantial parameter reduction, it achieves even higher performance on ARC-AGI-1 (Table~\ref{tab:full-cose-ablation}, Table~\ref{tab:full-cose-ablation-no-rearc}) and ARC-AGI-2 (Table~\ref{tab:component-grid}). When gating is enabled, the compositional branch controls how much the residual memory contributes to the task embedding. These results demonstrate a substantial improvement in the parameter efficiency of task embeddings; Appendix~\ref{app:pareto} visualizes the improved accuracy--parameter frontier.

\begin{table}[t]
  \centering
  \caption{\textbf{Improving the task-memory accuracy--parameter tradeoff.} \method{} uses $1/15$ as many task-memory parameters as the full table while improving pass@1 and pass@2. All rows use the same recurrent backbone, training data, and optimization setup, varying only the task-memory module. While the low-rank residual and compositional branch are less effective, their combination yields superior performance. Repeated rows report mean $\pm$ sample standard deviation; the remaining architectural variants are single runs.}
  \label{tab:full-cose-ablation}
  \small
  \resizebox{\linewidth}{!}{%
  \smalluncertainties
  \begin{tabular}{lrccc}
    \toprule
    Setup & Embedding params & pass@1 & pass@2 & pass@1000\\
    \midrule
    Full task table                         & 448.9M & $73.25 \pm 0.74$ & $78.50 \pm 1.03$ & $91.38 \pm 0.65$\\
    Rank-32 task table only                 & 28.1M  & 70.62 & 76.62 & 88.88\\
    Compositional branch only               & 1.5M   & 73.50 & 79.25 & 89.00\\
    \method{} (rank-32 residual)            & 29.6M  & $74.29 \pm 0.75$ & $80.71 \pm 0.51$ & $90.62 \pm 0.50$\\
    \method{} (gated $r{=}32$)              & 29.9M  & $74.59 \pm 0.74$ & $80.00 \pm 0.70$ & $90.94 \pm 1.10$\\
    \method{} (rank-512 residual)           & 450.4M & \textbf{77.75} & \textbf{81.50} & \textbf{92.12}\\
    \bottomrule
  \end{tabular}%
  }
\end{table}

\section{The \sysname{} system}
\label{sec:system}

Building on top of our compact structured task memory, we further combine matched synthetic augmentation, tuned recurrent-backbone design choices, and inference-time aggregation to the complete \sysname{} system. Table~\ref{tab:overall-arc-performance} reports the resulting full-system comparison.

\begin{table}[ht]
  \centering
  \caption[Public-evaluation accuracy of ARC systems trained from scratch.]{\textbf{Public-evaluation accuracy of ARC systems trained from scratch.}\protect\footnotemark}
  \xdef\arcBaselineSourcesFootnote{\thefootnote}
  \xdef\arcBaselineSourcesAnchor{Hfootnote.\theHfootnote}
  \label{tab:overall-arc-performance}
  \begin{tabular}{lcrrcccc}
    \toprule
                                   & Synthetic & \multicolumn{2}{c}{Parameters} & \multicolumn{2}{c}{ARC-AGI-1} & \multicolumn{2}{c}{ARC-AGI-2}\\
    \cmidrule(lr){3-4}\cmidrule(lr){5-6}\cmidrule(lr){7-8}
    Method                         & data & Backbone & ARC-1 all & pass@1 & pass@2 & pass@1 & pass@2\\
    \midrule
    HRM~\citep{wang2025hrm}        & N/A & 27M  & 475M & 34.4 & 40.3 & --   & 5.0\\
    TRM~\citep{jolicoeur2025trm}   & N/A & 7M   & 455M & 40.0 & 44.6 & 4.6  & 7.8\\
    URM~\citep{gao2025urm}         & N/A & 14M  & 462M & 53.8 & 61.8 & 16.0 & 18.2\\
    VARC~\citep{hu2025varc}        & Re-ARC & 73M  & 73M  & 54.5 & 60.5 & 9.44 & 11.11\\
    \midrule
    \textbf{\sysname{}} & Re-ARC(2)\footnotemark & 14M & \textbf{44M} & \textbf{78.6} & \textbf{84.0} & \textbf{39.3} & \textbf{46.7}\\
    \bottomrule
  \end{tabular}
\end{table}
\footnotetext[\arcBaselineSourcesFootnote]{\hypertarget{\arcBaselineSourcesAnchor}{}Baseline accuracies are from papers or project webpages; URM pass@2 is our reproduction using the authors' code.}
\footnotetext{\sysname{} uses Re-ARC on ARC-AGI-1, as in VARC; and our synthetic Re-ARC2 dataset (Section~\ref{sec:synthetic-data}) for ARC-AGI-2.}

\subsection{Programmatic data augmentation}
\label{sec:synthetic-data}

The Re-ARC dataset is a collection of synthetic augmentations for ARC-AGI-1. It provides 1{,}000 additional input-output pairs per training task, generated programmatically according to the underlying puzzle rule. However, the generation process is not fully automated, as the rule implementations are manually written by human.~\citep{hodel2024rearc}. While dihedral transformations and color permutations primarily encourage the model to learn positional and token invariance, Re-ARC augmentations provide stronger puzzle-specific priors that better bootstrap the model’s learning process during training.

Unlike standalone synthetic datasets, Re-ARC does not introduce additional priors beyond the scope of the original training distribution. The augmentations are applied only to the training split. Although comparisons with and without Re-ARC are therefore not strictly apples-to-apples, these two properties ensure that no information leaks into the evaluation and substantially reduce concerns regarding fairness. As shown in Table~\ref{tab:arc1-rearc-data-ablations}, incorporating Re-ARC leads to a notable performance improvement on ARC-AGI-1. Despite the substantial increase in dataset size introduced by Re-ARC and its augmentations, \method{} still maintains a performance advantage while using significantly fewer parameters.

Given the performance gains observed on ARC-AGI-1, it is natural to extend the same idea to ARC-AGI-2. Since ARC-AGI-2 does not have a public Re-ARC-style dataset, we curated a \emph{synthetic Re-ARC2} dataset for its 1000 training tasks. The objective is identical to that of Re-ARC: generating additional examples derived from the official task rules without introducing new task families. To construct this dataset, we designed an agentic workflow with a coding agent (Codex with GPT-5.4 xhigh) to generate task-specific data generator programs, using the same DSL (domain-specific language) defined in Re-ARC. The generated outputs were then verified both programmatically, through manual visual inspection, and a second-round audit by a model from a different family; flagged cases were manually reviewed, corrected where necessary, and re-audited.

On ARC-AGI-2, we compare the performance of different synthetic augmentation datasets under the same network architecture, and our synthetic Re-ARC2 demonstrates a clear advantage (Table~\ref{tab:arc2-rearc-data-ablations}). The two ingredients also interact super-additively: \method{}'s advantage over the full table grows from $+3.2$\% pass@2 without synthetic data to $+9.0$\% under Re-ARC2, consistent with compositional memory exploiting the richer rule-preserving coverage. Generation details and a quantitative audit are provided in Appendix~\ref{app:rearc2}. 

\begin{table}[t]
  \centering
  \caption{\textbf{Interaction between synthetic data and task memory on ARC-AGI-1.} Rows use the same backbone, training schedule, and default evaluator. Re-ARC gives large gains over the no-synthetic baseline, and \method{} retains those gains with compact task memory.}

  \label{tab:arc1-rearc-data-ablations}
  \smalluncertainties
  \begin{tabular}{lcccc}
    \toprule
    Synthetic data & Task embedding & pass@1 & pass@2 & pass@1000\\
    \midrule
    None  (baseline)                & full table  & $54.16 \pm 1.91$ & $59.97 \pm 1.77$ & $80.41 \pm 1.67$\\
    Re-ARC (aug=10)               & full table   & 68.0 & 72.5 & 84.63\\
    Re-ARC (aug=10)               & \method{}   & 70.88 & 76.38 & 86.25\\
    Re-ARC (aug=100)              & \method{}   & $\mathbf{74.29} \pm 0.75$ & $\mathbf{80.71} \pm 0.51$ & $\mathbf{90.62} \pm 0.50$\\
    \bottomrule
  \end{tabular}
\end{table}

\begin{table}[t]
  \centering
  \caption{\textbf{Matched synthetic data is the largest ARC-AGI-2 driver.} Rows use the same backbone, optimizer, training schedule, and default evaluator. ARC-AGI-1 Re-ARC tests cross-benchmark transfer; Re-ARC2 tests task-matched augmentation at increasing scale.}
  \label{tab:arc2-rearc-data-ablations}
  \small
  \resizebox{\linewidth}{!}{%
  \smalluncertainties
  \begin{tabular}{lcccc}
    \toprule
    Synthetic data & Task embedding & pass@1 & pass@2 & pass@1000\\
    \midrule
    None  (baseline)                & full table  & $13.16 \pm 0.65$ & $15.42 \pm 1.51$ & $30.00 \pm 2.91$\\
    None                            & \method{}   & $13.96 \pm 1.12$ & $18.61 \pm 1.63$ & $31.22 \pm 2.90$\\
    Re-ARC (ARC-AGI-1) (aug=100)  & \method{}   & 23.89 & 28.47 & 46.39\\
    Synthetic Re-ARC2 (aug=10)    & \method{}   & 30.69 & 35.97 & 45.97\\
    Synthetic Re-ARC2 (aug=100)   & full table  & $23.61 \pm 3.52$ & $28.23 \pm 3.58$ & $49.58 \pm 2.59$\\
    Synthetic Re-ARC2 (aug=100)   & \method{}   & $\mathbf{30.97} \pm 1.52$ & $\mathbf{37.22} \pm 1.77$ & $\mathbf{54.48} \pm 1.89$\\
    \bottomrule
  \end{tabular}%
  }
\end{table}

\subsection{The scaling tradeoffs for the recurrent backbone}
\label{sec:backbone-scaling}

The XRM backbone repeats a block of $U$ unique Transformer layers $R$ times and propagates gradients through a learning horizon $H$. In two ARC-AGI-1 scaling studies, we find that recurrent depth should be balanced with learning horizon. At matched forward and backward FLOPs, too many unique layers overfit while too few provide insufficient capacity. At fixed forward configuration, intermediate learning horizons outperform both very short and full horizons. We use the resulting balanced configuration in \sysname{}. The scaling methodology, visualization, and complete results are provided in Appendix~\ref{app:scaling-table}.

\subsection{Inference-time aggregation}
\label{sec:ttc-main}

The XRM evaluation protocol already aggregates votes across augmented puzzle views and training checkpoints. We strengthen this evaluator with an exponentially decayed replay of recent checkpoints (half-life of 10) and by ensembling outer refinement loops of 16 and 24 iterations. At the standard training budgets, the replay evaluator improves pass@2 from 79.62 to 81.50 on ARC-AGI-1 and from 38.47 to 39.30 on ARC-AGI-2 (Appendix~\ref{app:ttc}). These gains diminish as extended training approaches convergence, so the results in Table~\ref{tab:overall-arc-performance} use the standard protocol.

\FloatBarrier

\section{Generalization and continual learning}
\label{sec:generalization}

The evidence so far is specific to ARC and the XRM evaluation protocol. This raises two questions: Does the structured memory design help a trained model adapt to unseen tasks? And do its benefits generalize to other domains where memory keys share a common structure?

We first study continual learning by training ARC-AGI-1 models on unseen ARC-AGI-2 tasks. \method{} learns new puzzles efficiently through small task-specific memories without forgetting. We then adapt structured memory on a cellular-automata reasoning benchmark and on language modeling. Both domains show an improved accuracy--parameter trade-off, but the two branches play different roles: composition enables transfer to unseen rule--horizon combinations in cellular automata, while in language modeling, it adds little on its own but makes the lookup memory notably more effective.

\begin{figure}[!h]
  \centering
  \begin{minipage}[t]{0.49\linewidth}\centering
    \includegraphics[width=\linewidth]{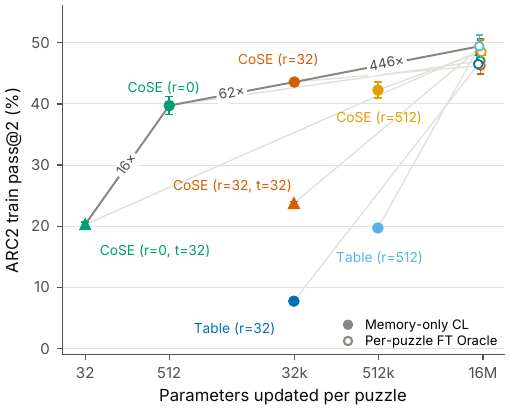}\\
    \footnotesize (a) memory-only continual learning vs. per-task adaptation
  \end{minipage}\hfill
  \begin{minipage}[t]{0.49\linewidth}\centering
    \includegraphics[width=\linewidth]{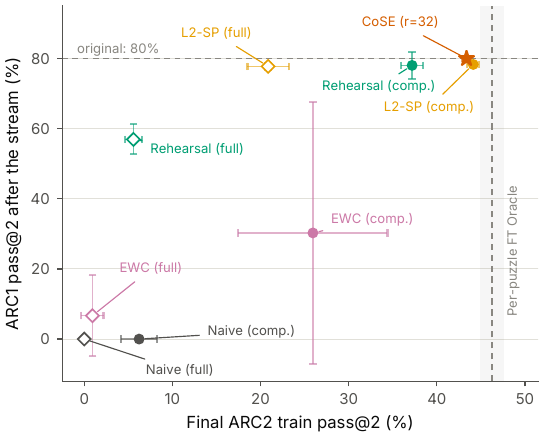}\\
    \footnotesize (b) continual-learning trade-off
  \end{minipage}
  \caption{\textbf{Frozen-backbone \method{} learns unseen tasks sequentially without forgetting.}
  (a) Composition-only \method{} ($r=0$) reaches $39.73\!\pm\!1.48$\% pass@2 with 512 trained parameters per puzzle; adding rank-32 residuals raises accuracy to $43.58\!\pm\!0.58$\% with 31.8k extra parameters. Reducing the compositional task embedding to 32 parameters per puzzle still reaches $20.30\!\pm\!0.44$\%.
  (b) Memory-only $r=32$ \method{} preserves 80.00\% ARC-AGI-1 accuracy. Composition-scoped L2-SP gains only 0.76\%  ARC-AGI-2 pass@2 ($44.13$\% versus $43.37$\%) at $35\times$ the updated parameters, while ARC-AGI-1 accuracy falls to 78.33\%.}
  \label{fig:cose-cl}
\end{figure}

\paragraph{Per-puzzle fine-tuning and continual adaptation.}

With only 512 trainable parameters per puzzle, \method{} adapts to unseen puzzles without forgetting and doubles the accuracy of the full table baseline, which updates about 1{,}000$\times$ as many parameters. Adding the rank-32 residual table raises accuracy further, still at $1/16$ the full table's parameter cost. From ARC-AGI-1 checkpoints, we adapt sequentially using each puzzle's demonstrations, updating only new memory rows while shared weights and earlier rows stay fixed. On 233 unseen ARC-AGI-2 training and 114 public-evaluation puzzles, \method{} ($r=32$) reaches 43.58\% and 12.19\% pass@2, versus 19.73\% and 4.47\% for the full table ($r=512$). Composition-only \method{} ($r=0$) reaches 39.73\% and 10.76\% (Figure~\ref{fig:cose-cl}(a)). 

The full table baseline does not lack capacity: when the whole model is fine-tuned on each puzzle independently (Per-puzzle FT Oracle), it reaches comparable accuracy to \method{}, indicating that its memory architecture limits efficient adaptation. Our composition module is also a better target for standard continual-learning methods. Rehearsal~\citep{robins1995rehearsal}, L2-SP~\citep{xuhong2018l2sp}, and EWC~\citep{kirkpatrick2017ewc} all perform better when shared updates are restricted to the composition module than the full model (Figure~\ref{fig:cose-cl}(b)). We omit the auxiliary halting loss for all methods so that each receives the same number of training updates; Appendix~\ref{app:cose-cl} gives protocols and ablations.

\begin{figure}[!h]
  \centering
  \begin{minipage}[t]{0.49\linewidth}\centering
    \includegraphics[width=\linewidth]{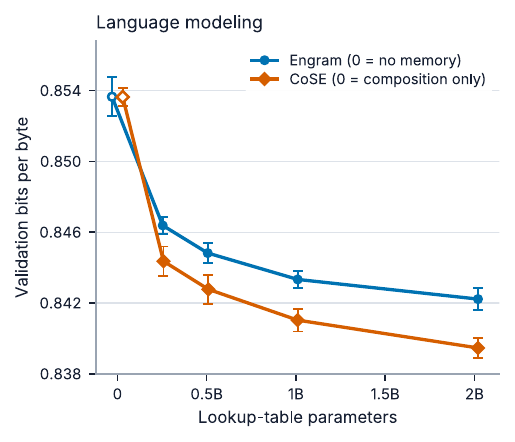}\\
    \footnotesize (a) hashed n-gram memory in nanochat
  \end{minipage}\hfill
  \begin{minipage}[t]{0.49\linewidth}\centering
    \includegraphics[width=\linewidth]{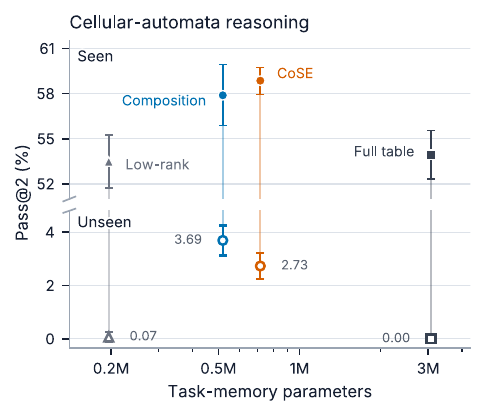}\\
    \footnotesize (b) rule--horizon transfer in CAR
  \end{minipage}
  \caption{\textbf{Structured composition transfers across domains and task keys.}
  (a) On a compute-optimal nanochat d12 model, adding the \method{} compositional path to Engram lowers validation bits per byte at every table size, matching Engram at twice the capacity (circles: Engram; diamonds: \method{}).
  (b) On cellular-automata reasoning (CAR), composition enables transfer from 1{,}500 trained rule--horizon combinations to 4{,}500 unseen combinations, where independent lookup is nearly zero. \method{} is strongest on seen combinations, while removing its  residual table improves unseen transfer.}
  \label{fig:cose-generalization}
\end{figure}

\paragraph{Cellular-automata reasoning.}

Cellular-Automata Reasoning (CAR) is a synthetic 2D-grid puzzle benchmark inspired by Conway's Game of Life~\citep{gardner1970fantastic}. Each task specifies a rule that updates every cell simultaneously from its local neighborhood; given random $16\times16$ input grids, the model must predict the exact grid after $T\in\{1,2,4,8\}$ updates. Multi-step prediction is challenging because local errors compound across the rollout. CAR contains 1{,}500 rules, and we train each rule at only one horizon before evaluating on both the 1{,}500 observed rule--horizon combinations and the remaining 4{,}500 unseen combinations. Using a rule--horizon--dihedral memory key, three-factor \method{} achieves the strongest seen accuracy, improving pass@2 from $53.94$\% for a full table to $58.86$\% while using $23$\% as many task-memory parameters. On unseen combinations, the full table cannot transfer ($0$\% pass@2), whereas structured composition reaches $3.69$\%. The composition-only model outperforms \method{}, showing that the residual improves fitting of observed combinations but can slightly weaken combinatorial transfer.

\paragraph{Language modeling.}
We test the structured memory in language-model pretraining, a setting with no explicit task identities. Engram uses hashed 2- and 3-gram contexts to index a large auxiliary lookup table; so the token context itself serves as the structured memory key. We implement Engram in nanochat and augment it with a shared FiLM module composed from the current and preceding two token embeddings. On a compute-optimal 286M-parameter d12 model ($1.19$B pretraining tokens), we compare Engram and \method{} at four memory capacities ranging from $0.25$B to $2.02$B parameters. \method{} achieves lower validation bits per byte at every capacity, demonstrating a consistently better accuracy--parameter trade-off. The advantage persists on a larger d20 model, although the margin narrows as the amount of training data increases. Appendix~\ref{app:generalization} provides the complete protocols, scaling and results.

\section{Limitations}

We follow the prior XRM training protocol, which uses demonstration pairs from public evaluation tasks but never their test outputs. The ARC Prize team's independent analysis concluded that this  ``does not imply any data leakage''~\citep{arcprize2025hrmanalysis}. Section~\ref{sec:generalization} further evaluates \method{} on unseen puzzles using only their demonstrations---the data-access pattern of in-context learning. These accuracies are not directly comparable to our final results because this experiment transfers an ARC-AGI-1-trained model to ARC-AGI-2.

We repeat key ablations when feasible, but the cost of XRM training limits some experiments to a single run. We prioritize breadth across architectural variants over repeated identical configurations. Exploratory runs can also be terminated early when their performance is clearly outside the relevant range.

\section{Discussion}

\sysname{} achieves strong ARC-AGI performance while substantially improving parameter efficiency. Across broader domains, our results suggest that the compositional branch is most effective when memory keys compose into meaningful yet diversified local contexts, as with rule--horizon pairs or local n-grams. Its low-rank residual is useful when instance-specific memory remains beneficial but a full-width table is overparameterized. More broadly, recurrent reasoning is appealing in data-scarce, reasoning-intensive regimes: iterative refinement increases computation without increasing backbone capacity, while \method{} reduces redundant memory capacity. This combination may benefit structured scientific problems such as inverse problems, state estimation, and learned iterative solvers, where data are limited and repeated refinement is natural.

\begin{ack}
This work was supported by Laboratory Directed Research and Development (LDRD) funding from Argonne National Laboratory, provided by the Director, Office of Science, U.S. Department of Energy, under Contract No. DE-AC02-06CH11357, and by the University of Chicago Joint Task Force Initiative. This research used resources of the Argonne Leadership Computing Facility, a U.S. Department of Energy Office of Science user facility at Argonne National Laboratory.
The authors declare no competing interests.
\end{ack}

\bibliographystyle{unsrtnat}
\bibliography{references}

@article{chollet2019measure,
  title={On the measure of intelligence},
  author={Chollet, Fran{\c{c}}ois},
  journal={arXiv preprint arXiv:1911.01547},
  year={2019}
}

@article{chollet2025arc2,
  title={{ARC-AGI-2}: A New Challenge for Frontier {AI} Reasoning Systems},
  author={Chollet, Fran{\c{c}}ois and Knoop, Mike and Kamradt, Gregory and Landers, Bryan and Pinkard, Henry},
  journal={arXiv preprint arXiv:2505.11831},
  year={2025}
}

@article{wang2025hrm,
  title={Hierarchical reasoning model},
  author={Wang, Guan and Li, Jin and Sun, Yuhao and Chen, Xing and Liu, Changling and Wu, Yue and Lu, Meng and Song, Sen and Yadkori, Yasin Abbasi},
  journal={arXiv preprint arXiv:2506.21734},
  year={2025}
}

@article{jolicoeur2025trm,
  title={Less is more: Recursive reasoning with tiny networks},
  author={Jolicoeur-Martineau, Alexia},
  journal={arXiv preprint arXiv:2510.04871},
  year={2025}
}

@article{gao2025urm,
  title={Universal Reasoning Model},
  author={Gao, Zitian and Chen, Lynx and Xiao, Yihao and Xing, He and Tao, Ran and Luo, Haoming and Zhou, Joey and Dai, Bryan},
  journal={arXiv preprint arXiv:2512.14693},
  year={2025}
}

@article{hu2025varc,
  title={{ARC} Is a Vision Problem!},
  author={Hu, Keya and Cy, Ali and Qiu, Linlu and Ding, Xiaoman Delores and Wang, Runqian and Zhu, Yeyin Eva and Andreas, Jacob and He, Kaiming},
  journal={arXiv preprint arXiv:2511.14761},
  year={2025}
}

@article{cheng2026engram,
  title={Conditional memory via scalable lookup: A new axis of sparsity for large language models},
  author={Cheng, Xin and Zeng, Wangding and Dai, Damai and Chen, Qinyu and Wang, Bingxuan and Xie, Zhenda and Huang, Kezhao and Yu, Xingkai and Hao, Zhewen and Li, Yukun and others},
  journal={arXiv preprint arXiv:2601.07372},
  year={2026}
}

@article{dehghani2019universal,
  title={Universal transformers},
  author={Dehghani, Mostafa and Gouws, Stephan and Vinyals, Oriol and Uszkoreit, Jakob and Kaiser, {\L}ukasz},
  journal={arXiv preprint arXiv:1807.03819},
  year={2018}
}

@inproceedings{perez2018film,
  title={{FiLM}: Visual reasoning with a general conditioning layer},
  author={Perez, Ethan and Strub, Florian and de Vries, Harm and Dumoulin, Vincent and Courville, Aaron},
  booktitle={Proceedings of the AAAI conference on artificial intelligence},
  volume={32},
  year={2018}
}

@article{hodel2024rearc,
  title={Addressing the abstraction and reasoning corpus via procedural example generation},
  author={Hodel, Michael},
  journal={arXiv preprint arXiv:2404.07353},
  year={2024}
}

@article{moskvichev2023conceptarc,
  title={The {ConceptARC} benchmark: Evaluating understanding and generalization in the {ARC} domain},
  author={Moskvichev, Arseny and Odouard, Victor Vikram and Mitchell, Melanie},
  journal={arXiv preprint arXiv:2305.07141},
  year={2023}
}

@article{architects2024,
  title={Product of experts with {LLMs}: Boosting performance on {ARC} is a matter of perspective},
  author={Franzen, Daniel and Disselhoff, Jan and Hartmann, David},
  journal={arXiv preprint arXiv:2505.07859},
  year={2025}
}

@misc{jordan2024muon,
  author       = {Keller Jordan and Yuchen Jin and Vlado Boza and Jiacheng You and
                  Franz Cesista and Laker Newhouse and Jeremy Bernstein},
  title        = {Muon: An optimizer for hidden layers in neural networks},
  year         = {2024},
  url          = {https://kellerjordan.github.io/posts/muon/}
}

@inproceedings{hashembed,
  title={Hash embeddings for efficient word representations},
  author={Tito Svenstrup, Dan and Hansen, Jonas and Winther, Ole},
  booktitle={Advances in Neural Information Processing Systems},
  volume={30},
  year={2017}
}

@misc{arcprize2025hrmanalysis,
  author       = {{ARC Prize Team}},
  title        = {The Hidden Drivers of {HRM}'s Performance on {ARC-AGI}},
  year         = {2025},
  month        = aug,
  howpublished = {\url{https://arcprize.org/blog/hrm-analysis}},
  note         = {Published August 15, 2025. Accessed May 5, 2026}
}

@article{gardner1970fantastic,
  author  = {Gardner, Martin},
  title   = {Mathematical Games: The Fantastic Combinations of {John Conway}'s New Solitaire Game ``Life''},
  journal = {Scientific American},
  volume  = {223},
  number  = {4},
  pages   = {120--123},
  year    = {1970}
}

@misc{karpathy2025nanochat,
  author       = {Karpathy, Andrej},
  title        = {nanochat: The Best {ChatGPT} That \$100 Can Buy},
  year         = {2025},
  howpublished = {\url{https://github.com/karpathy/nanochat}}
}

@article{kirkpatrick2017ewc,
  author  = {Kirkpatrick, James and Pascanu, Razvan and Rabinowitz, Neil and Veness, Joel and Desjardins, Guillaume and Rusu, Andrei A. and Milan, Kieran and Quan, John and Ramalho, Tiago and Grabska-Barwinska, Agnieszka and Hassabis, Demis and Clopath, Claudia and Kumaran, Dharshan and Hadsell, Raia},
  title   = {Overcoming Catastrophic Forgetting in Neural Networks},
  journal = {Proceedings of the National Academy of Sciences},
  volume  = {114},
  number  = {13},
  pages   = {3521--3526},
  year    = {2017}
}

@inproceedings{xuhong2018l2sp,
  author    = {Li, Xuhong and Grandvalet, Yves and Davoine, Franck},
  title     = {Explicit Inductive Bias for Transfer Learning with Convolutional Networks},
  booktitle = {International Conference on Machine Learning},
  pages     = {2825--2834},
  year      = {2018}
}

@article{robins1995rehearsal,
  author  = {Robins, Anthony},
  title   = {Catastrophic Forgetting, Rehearsal and Pseudorehearsal},
  journal = {Connection Science},
  volume  = {7},
  number  = {2},
  pages   = {123--146},
  year    = {1995}
}

\appendix
\section{Task-conditioned memory: additional details and ablations}
\label{app:task-memory}

This appendix collects the material supporting the task-memory study in Section~\ref{sec:task-memory}. We first present the full \method{} module (Appendix~\ref{app:cose-module}), its parameter accounting (Appendix~\ref{app:param-count}), and the complete repeated-run results behind Table~\ref{tab:component-grid} (Appendix~\ref{app:arc-repeats}). We then establish that the task-memory advantage persists without synthetic data (Appendix~\ref{app:cose-no-rearc}), show the resulting accuracy--parameter frontier (Appendix~\ref{app:pareto}), and report the design ablations: alternative composition operators (Appendix~\ref{app:composition-ablations}) and residual variants (Appendix~\ref{app:cose-ablations}).

\subsection{Full \method{} module diagram}
\label{app:cose-module}

\begin{figure}[ht]
  \centering
  \includegraphics[width=0.78\linewidth]{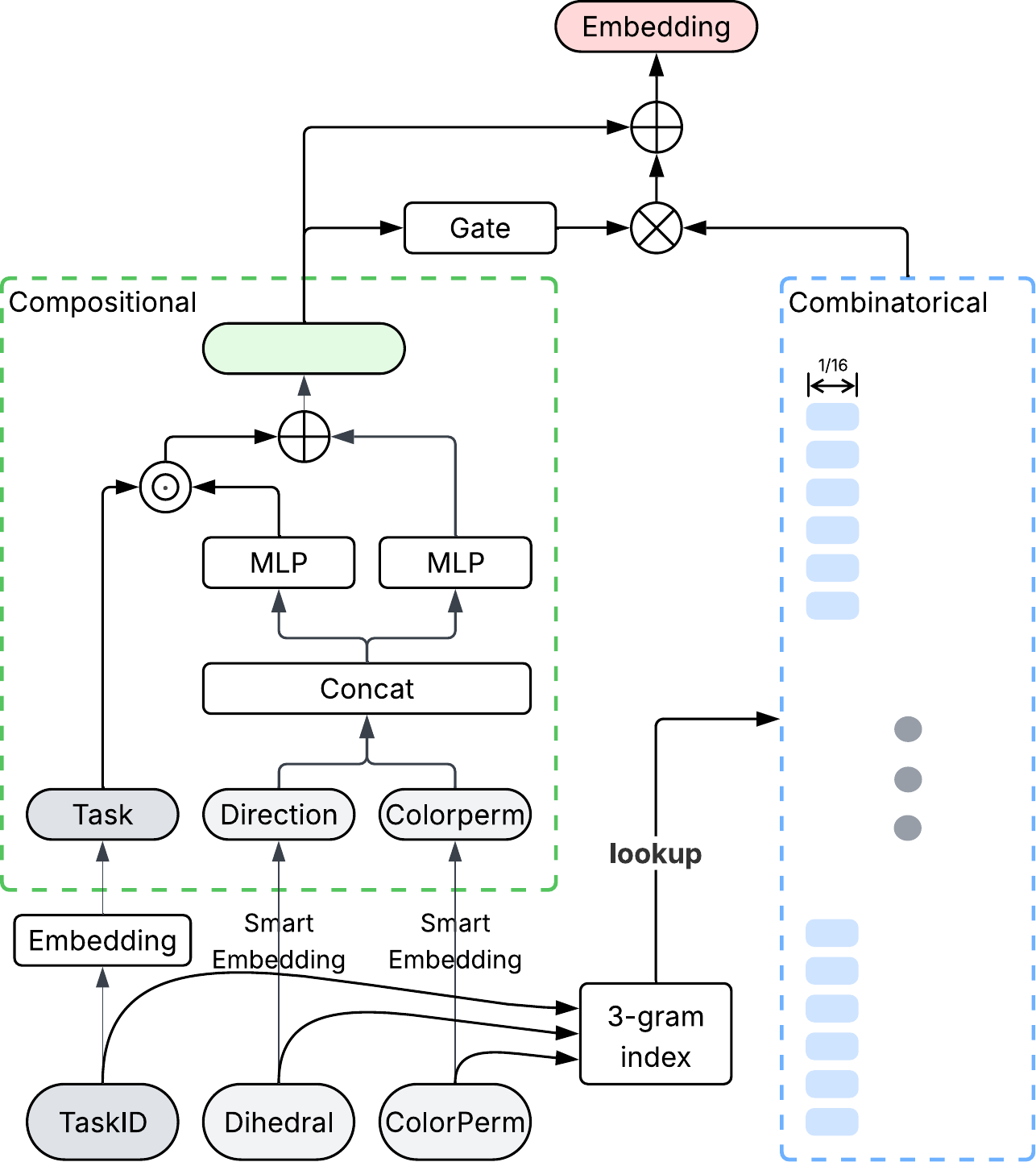}
  \caption{\textbf{Expanded data flow for the \method{} task embedding.}
  Puzzle identity and dihedral transform are read from small embedding tables; color permutation is represented by structure-preserving embeddings that permute color-slot sub-vectors.
  The augmentation embeddings modulate the puzzle embedding through a FiLM-like operator to form the compositional branch.
  The original per-instance index retrieves a rank-$32$ row, which is 1/16 of the original width; this row is up-projected, gated by the compositional embedding, and added back to produce the final task embedding.}
  \label{fig:cose-detail}
\end{figure}

\FloatBarrier

\subsection{Parameter accounting details}
\label{app:param-count}

Table~\ref{tab:embedding-params} summarizes the task-memory footprint across dataset setups. Tables~\ref{tab:param-decomp-arc1}--\ref{tab:param-arc2-exact} provide the component-level formulas and exact counts behind these totals.

\begin{table}[t]
  \centering
  \caption{\textbf{Task-memory footprint in learnable parameters.} \method{} reduces learned task-memory parameters by 94\% across ARC-AGI dataset setups. \textit{Rows} counts combinatorial puzzle-augmentation instances. \textit{Task memory} is the full task table for stock models and the compositional branch plus low-rank residual table for \method{}.}
  \label{tab:embedding-params}
  \small
  \begin{tabular}{llrrr}
    \toprule
    Setting & Model & Rows & Task memory & Total\\
    \midrule
    ARC-AGI-1 & URM, full table & 0.876M & 448.3M & 462.0M\\
    ARC-AGI-1 + Re-ARC & URM, full table & 0.877M & 448.9M & 462.5M\\
    ARC-AGI-1 + Re-ARC & \sysname{} ($r{=}32$ + gate) & 0.877M & 29.9M & 43.5M\\
    \midrule
    ARC-AGI-2 & URM, full table & 1.191M & 609.6M & 623.3M\\
    ARC-AGI-2 & \sysname{} ($r{=}32$) & 1.191M & 39.8M & 53.5M\\
    ARC-AGI-2 + Re-ARC2 & \sysname{} ($r{=}32$) & 1.289M & 43.0M & 56.6M\\
    \bottomrule
  \end{tabular}
\end{table}

\begin{table}[ht]
  \centering
  \caption{\textbf{Component-level parameter accounting for ARC-AGI-1 with Re-ARC augmentation.} The table reports the measured recurrent backbone and the analytic counts for the full task table, \method{} compositional branch, rank-32 residual table, up-projection, and optional gate.}
  \label{tab:param-decomp-arc1}
  \small
  \begin{tabular}{llr}
    \toprule
    Component & Count & Parameters\\
    \midrule
    URM backbone & measured & $13{,}663{,}233$\\
    Full instance table & $Md$ & $448{,}872{,}960$\\
    \method{} compositional branch & $(P+9)d+2(2d^2+d)$ & $1{,}545{,}728$\\
    \method{} residual table & $Mr$ & $28{,}054{,}560$\\
    \method{} residual up-projection & $rd$ & $16{,}384$\\
    \method{} optional gate & $d^2+d$ & $262{,}656$\\
    \midrule
    Stock URM + full table & measured & $462{,}536{,}193$\\
    \sysname{} without gate & measured & $43{,}279{,}905$\\
    \sysname{} with gate & measured & $43{,}542{,}561$\\
    \bottomrule
  \end{tabular}
\end{table}

\begin{table}[ht]
  \centering
  \caption{\textbf{Exact ARC-AGI-1 parameter counts across task-memory designs.} Other memory includes the compositional branch, any residual up-projection, and the optional gate. Full tables have width 512; rank-$r$ tables have width $r$. Composition-only has no instance table.}
  \label{tab:param-arc1-exact}
  \small
  \setlength{\tabcolsep}{3pt}
  \begin{tabular}{@{}lrrrr@{}}
    \toprule
    Design & Backbone & Other memory & Instance table & Total\\
    \midrule
    \multicolumn{5}{@{}l}{\textit{ARC-AGI-1: 875,617 instance rows}}\\
    \addlinespace[2pt]
    TRM, full table & $6{,}830{,}082$ & $0$ & $448{,}315{,}904$ & $455{,}145{,}986$\\
    URM, full table & $13{,}663{,}746$ & $0$ & $448{,}315{,}904$ & $461{,}979{,}650$\\
    \midrule
    \multicolumn{5}{@{}l}{\textit{ARC-AGI-1 + Re-ARC: 876,705 instance rows when present}}\\
    \addlinespace[2pt]
    Composition only & $13{,}663{,}233$ & $1{,}545{,}728$ & $0$ & $15{,}208{,}961$\\
    URM, full table & $13{,}663{,}233$ & $0$ & $448{,}872{,}960$ & $462{,}536{,}193$\\
    Rank-32 table only & $13{,}663{,}233$ & $16{,}384$ & $28{,}054{,}560$ & $41{,}734{,}177$\\
    \method{}, $r=32$ & $13{,}663{,}233$ & $1{,}562{,}112$ & $28{,}054{,}560$ & $43{,}279{,}905$\\
    \method{}, $r=32$ + gate & $13{,}663{,}233$ & $1{,}824{,}768$ & $28{,}054{,}560$ & $43{,}542{,}561$\\
    \method{}, $r=512$ & $13{,}663{,}233$ & $1{,}545{,}728$ & $448{,}872{,}960$ & $464{,}081{,}921$\\
    \bottomrule
  \end{tabular}
\end{table}

\begin{table}[ht]
  \centering
  \caption{\textbf{Exact ARC-AGI-2 parameter counts.} Columns follow Table~\ref{tab:param-arc1-exact}. Full tables have width 512, and \method{} residual tables have width 32. Group headings specify the dataset and number of instance rows.}
  \label{tab:param-arc2-exact}
  \small
  \setlength{\tabcolsep}{3pt}
  \begin{tabular}{@{}lrrrr@{}}
    \toprule
    Design & Backbone & Other memory & Instance table & Total\\
    \midrule
    \multicolumn{5}{@{}l}{\textit{ARC-AGI-2: 1,190,624 instance rows}}\\
    \addlinespace[2pt]
    URM, full table & $13{,}663{,}746$ & $0$ & $609{,}599{,}488$ & $623{,}263{,}234$\\
    \method{}, $r=32$ & $13{,}663{,}233$ & $1{,}725{,}952$ & $38{,}099{,}968$ & $53{,}489{,}153$\\
    \midrule
    \multicolumn{5}{@{}l}{\textit{ARC-AGI-2 + Re-ARC2: 1,289,151 instance rows}}\\
    \addlinespace[2pt]
    \method{}, $r=32$ & $13{,}663{,}233$ & $1{,}725{,}952$ & $41{,}252{,}832$ & $56{,}642{,}017$\\
    \bottomrule
  \end{tabular}
\end{table}

\FloatBarrier

\subsection{Repeated ARC task-memory and synthetic-data ablations}
\label{app:arc-repeats}

Table~\ref{tab:arc-repeat-ablation} expands the main component comparison (Table~\ref{tab:component-grid}) with pass@1, pass@2, and pass@1000 for both gated and ungated \method{}. Each setup has 3--4 repeated runs. 

\begin{table}[ht]
  \centering
  \caption{\textbf{Complete component ablation with repeated runs.} The comparison studies benchmark, matched synthetic data, and task-memory design. Bold values are the highest means within each benchmark--data block.}
  \label{tab:arc-repeat-ablation}
  \scriptsize
  \setlength{\tabcolsep}{3.5pt}
  \resizebox{\linewidth}{!}{%
  \smalluncertainties
  \begin{tabular}{lllrrrr}
    \toprule
    Benchmark & Synthetic data & Task memory & $n$ & pass@1 & pass@2 & pass@1000\\
    \midrule
    \multirow{6}{*}{ARC-AGI-1}
      & \multirow{3}{*}{None}
      & Full task table & 4 & $54.16 \pm 1.91$ & $59.97 \pm 1.77$ & $80.41 \pm 1.67$\\
      & & \method{} (ungated) & 3 & $56.46 \pm 1.51$ & $62.67 \pm 2.43$ & $79.96 \pm 1.81$\\
      & & \method{} (gated) & 4 & $\mathbf{57.41} \pm 1.82$ & $\mathbf{64.91} \pm 2.25$ & $\mathbf{81.41} \pm 1.65$\\
      \cmidrule(lr){2-7}
      & \multirow{3}{*}{Re-ARC}
      & Full task table & 4 & $73.25 \pm 0.74$ & $78.50 \pm 1.03$ & $\mathbf{91.38} \pm 0.65$\\
      & & \method{} (ungated) & 3 & $74.29 \pm 0.75$ & $\mathbf{80.71} \pm 0.51$ & $90.62 \pm 0.50$\\
      & & \method{} (gated) & 4 & $\mathbf{74.59} \pm 0.74$ & $80.00 \pm 0.70$ & $90.94 \pm 1.10$\\
    \midrule
    \multirow{6}{*}{ARC-AGI-2}
      & \multirow{3}{*}{None}
      & Full task table & 4 & $13.16 \pm 0.65$ & $15.42 \pm 1.51$ & $30.00 \pm 2.91$\\
      & & \method{} (ungated) & 4 & $\mathbf{13.96} \pm 1.12$ & $\mathbf{18.61} \pm 1.63$ & $31.22 \pm 2.90$\\
      & & \method{} (gated) & 3 & $13.80 \pm 1.04$ & $17.69 \pm 1.67$ & $\mathbf{32.22} \pm 1.10$\\
      \cmidrule(lr){2-7}
      & \multirow{3}{*}{Re-ARC2}
      & Full task table & 4 & $23.61 \pm 3.52$ & $28.23 \pm 3.58$ & $49.58 \pm 2.59$\\
      & & \method{} (ungated) & 4 & $\mathbf{30.97} \pm 1.52$ & $\mathbf{37.22} \pm 1.77$ & $\mathbf{54.48} \pm 1.89$\\
      & & \method{} (gated) & 3 & $28.84 \pm 3.99$ & $35.93 \pm 1.31$ & $53.94 \pm 3.46$\\
    \bottomrule
  \end{tabular}%
  }
\end{table}

\clearpage

\subsection{Task-memory tradeoff without Re-ARC dataset}
\label{app:cose-no-rearc}

Table~\ref{tab:full-cose-ablation-no-rearc} repeats the task-memory comparison of Table~\ref{tab:full-cose-ablation} without Re-ARC, matching the original XRM data setup. \method{}'s advantage over the full table is even larger in this setup.

\begin{table}[ht]
  \centering
  \caption{\textbf{Improving the task-memory accuracy--parameter tradeoff without Re-ARC dataset.} This table mirrors Table~\ref{tab:full-cose-ablation}, but omits Re-ARC from the ARC-AGI-1 training data, matching the original XRM setups. The recurrent backbone, training data, and optimization setup are fixed, varying only the task-memory module. All models are trained for at most 600k steps.}
  \label{tab:full-cose-ablation-no-rearc}
  \small
  \resizebox{\linewidth}{!}{%
  \smalluncertainties
  \begin{tabular}{lrccc}
    \toprule
    Setup & Embedding params & pass@1 & pass@2 & pass@1000\\
    \midrule
    Full task table                         & 448.3M & $54.16 \pm 1.91$ & $59.97 \pm 1.77$ & $80.41 \pm 1.67$\\
    Rank-32 task table only                 & 28.0M  & 52.88 & 58.00 & 79.25\\
    Compositional branch only               & 1.5M   & 47.38 & 51.50 & 70.12\\
    \method{} (rank-32 residual)            & 29.6M  & $56.46 \pm 1.51$ & $62.67 \pm 2.43$ & $79.96 \pm 1.81$\\
    \textbf{\method{} (gated $r{=}32$)}    & 29.8M  & $\mathbf{57.41} \pm 1.82$ & $\mathbf{64.91} \pm 2.25$ & $\mathbf{81.41} \pm 1.65$\\
    \bottomrule
  \end{tabular}%
  }
\end{table}

\FloatBarrier

\subsection{Pareto frontier visualization}
\label{app:pareto}

\begin{figure}[t]
  \centering
  \includegraphics[width=0.88\linewidth]{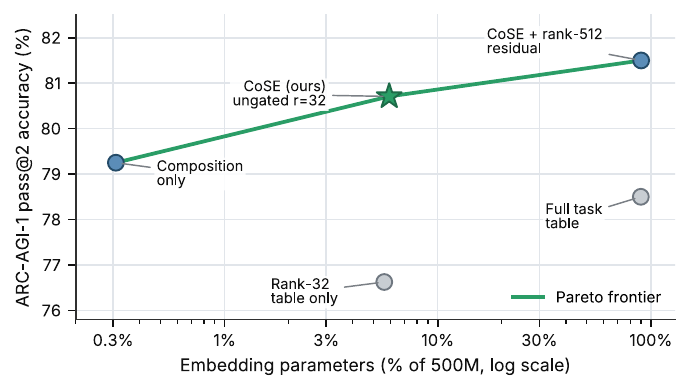}
  \caption{\textbf{\method{} improves the task-memory accuracy--parameter frontier.}
  All points use the same recurrent backbone, training data, and optimization setup, varying only the task-memory module.
  The compositional branch and low-rank residual are weaker on their own, but their combination exceeds the full table while using roughly $1/15$ task-memory parameters.}
  \label{fig:embedding-pareto}
\end{figure}

Figure~\ref{fig:embedding-pareto} plots the task-memory designs of Table~\ref{tab:full-cose-ablation} in the accuracy--parameter plane. \method{} does not trade accuracy for size: it improves on the frontier of the full task table while using about $1/15$ as many task-memory parameters.

\FloatBarrier

\subsection{Composition operator ablations}
\label{app:composition-ablations}

Table~\ref{tab:compo-ablations} compares operators for combining the puzzle, dihedral, and color embeddings in the ARC-AGI-1 training-split setting of Section~\ref{sec:task-memory}. Several operators improve over simple addition, but all remain below the full lookup table, motivating the per-instance residual in \method{}.

\begin{table}[ht]
  \centering
  \caption{\textbf{Pure compositional task conditioning does not fully recover the instance-specific capacity of the task table.} On ARC-AGI-1 training tasks, several operators for combining puzzle, dihedral, and color information improve over simple addition but remain below the full lookup baseline. This persistent gap motivates adding a small per-instance residual in \method{}. Values are percentage accuracies, reported as mean $\pm$ std when multiple recovered runs are available.}
  \label{tab:compo-ablations}
  \small
  \resizebox{\linewidth}{!}{%
  \smalluncertainties
  \begin{tabular}{lllcc}
    \toprule
    Dihedral Embedding & Color Embedding & Combination method & pass@2 & pass@1000\\
    \midrule
    Full task table & Full task table & Lookup & $49.75 \pm 2.50$ & $71.75 \pm 0.22$\\
    \midrule
    Lookup table & Slotperm  & FiLM-concat & $47.50 \pm 1.16$ & $63.25 \pm 2.01$\\
    Slotperm & Slotperm & FiLM-concat & $46.77 \pm 0.27$ & $64.50 \pm 1.59$\\
    Lookup table & Slotperm & Hypernetwork & 46.46 & 59.38\\
    Lookup table & Slotperm  & More expressive FiLM & $46.09 \pm 2.30$ & $62.00 \pm 0.88$\\
    Lookup table & Slotperm  & FiLM-concat, larger backbone & $45.58 \pm 0.71$ & $64.13 \pm 0.35$\\
    Lookup table & Slotperm  & FiLM-add & $45.21 \pm 3.23$ & $63.88 \pm 2.28$\\
    Lookup table & Slotperm & Mixture of experts & 44.79 & 60.25\\
    Lookup table & Slotperm  & FiLM-concat, 1024 hidden units & 44.58 & 63.04\\
    Lookup table & Slotperm  & Add & 44.21 & 63.50\\
    Lookup table & Slotperm  & FiLM-concat, 4 heads & 43.83 & 61.50\\
    \bottomrule
  \end{tabular}%
  }
\end{table}

\FloatBarrier

\subsection{\method{} residual ablations}
\label{app:cose-ablations}

Table~\ref{tab:cose-arc1training-ablations} ablates the residual branch in the same setting. Gated and ungated rank-32 residuals both surpass the full-table baseline at pass@2 while narrowing the pass@1000 gap left by pure composition; an Engram-style keyed residual is slightly weaker than direct per-instance indexing.

\begin{table}[ht]
  \centering
  \caption{\textbf{Residual-memory design ablation for compositional task conditioning on ARC-AGI-1 training tasks.} Starting from the compositional setup in Table~\ref{tab:compo-ablations}, we compare gated, ungated, and Engram-style rank-32 residuals under the same TRM backbone and training augmentation set. The full task-table row is included as the lookup baseline.}
  \label{tab:cose-arc1training-ablations}
  \small
  \resizebox{\linewidth}{!}{%
  \smalluncertainties
  \begin{tabular}{lllcc}
    \toprule
    Dihedral Embedding & Color Embedding & Combination/residual method & pass@2 & pass@1000\\
    \midrule
    Full task table & Full task table & Lookup & $49.75 \pm 2.50$ & $71.75 \pm 0.22$\\
    \midrule
    Slotperm & Slotperm & FiLM-concat + gated residual ($r=32$) & $51.83 \pm 0.76$ & $68.96 \pm 1.48$\\
    Slotperm & Slotperm & FiLM-concat + residual ($r=32$) & $51.25 \pm 0.94$ & $67.19 \pm 0.80$\\
    Slotperm & Slotperm & FiLM-concat + Engram residual ($r=32$) & 50.08 & 66.88\\
    \bottomrule
  \end{tabular}%
  }
\end{table}

\FloatBarrier

\section{Generalizing \method{} beyond ARC-AGI}
\label{app:generalization}

The ARC factorization in Equation~\ref{eq:factors} is task-specific. This appendix tests the general design principles behind \method{}: a structured key should be represented by a shared compositional path together with lookup capacity for key-specific exceptions. We first construct a synthetic reasoning benchmark whose factors and held-out combinations are known exactly, then test a different factorization  inside an autoregressive language model.

\subsection{Cellular-automata reasoning}
\label{app:car}

\paragraph{Benchmark.}
Cellular-Automata Reasoning (CAR) tasks ask the model to apply a 2D cellular-automaton rule for a specified number of steps $T$ on a grid. It contains 1{,}500 rules split evenly across Life-like, Generations, and Larger-than-Life rule families. For each rule, random $16\!\times\!16$ toroidal input grids with live-cell density sampled uniformly from $[0.1,0.5)$ are paired with the exact grid after $T\in\{1,2,4,8\}$ synchronous updates. We retain rules that are viable on at least 8 of 64 runs and require every output to contain both cell states and differ from its input. Every rule--horizon combination has 500 training examples and four held-out evaluation examples, using the same initial grids across horizons. In the \emph{single-horizon} split, one horizon is assigned to each rule, exactly balanced at 375 rules per horizon; the remaining 4{,}500 combinations test whether the model can apply a known rule to an untrained horizon. The \emph{all-horizons} split contains all 6{,}000 combinations.

\paragraph{Backbone and task memory.}
All models share a 1.710M-parameter recurrent backbone with hidden width 256, two unique Transformer layers, six recurrent applications, a three-step learning horizon, and four outer refinement steps. The structured memory key is rule identity $r$, rollout horizon $T$, and dihedral view $g$. The full table stores an independent 256-dimensional row for every $(r,T,g)$ combination; the low-rank table uses rank 16. To make the two structured variants explicit, let $Q_{r,T}$ be a joint rule--horizon embedding and let $R_r$, $H_T$, and $D_g$ be separate rule, horizon, and view embeddings. Their compositional memories are
\begin{equation}
  \begin{aligned}
    s^{(2)}_{r,T}&=Q_{r,T}, & s^{(3)}_{r,T}&=R_r+H_T,\\
    \gamma_g&=W_\gamma D_g+b_\gamma, & \beta_g&=W_\beta D_g+b_\beta,\\
    e^{(k)}_{r,T,g}&=(1+\gamma_g)\odot s^{(k)}_{r,T}+\beta_g,
      && k\in\{2,3\}.
  \end{aligned}
  \label{eq:car-factorization}
\end{equation}
So two-factor treats $(r,T)$ as one factor and $g$ as the other, while three-factor separates all three. The corresponding \method{} model adds $W_z z_{r,T,g}$ to $e^{(k)}_{r,T,g}$, where $z_{r,T,g}\in\mathbb{R}^{16}$ is a learned combination-specific residual; its reserved row is zero for an unseen $(r,T)$. Because $Q_{r,T}$ has no entry for such a pair, the two-factor design is evaluated only on observed rule--horizon combinations.

\paragraph{Training details.}
We train in bfloat16 with AdamW at dense learning rate $10^{-4}$, task-memory learning rate $10^{-2}$, $\beta=(0.9,0.95)$, weight decay 0.1, global batch 1{,}536, and 2{,}000 warmup steps followed by a constant rate. Training applies random dihedral views and places grids on a $30\!\times\!30$ canvas, at the origin with probability 0.2 and otherwise at a random translation. Single-horizon runs train for 500k updates to watch for emergent behaviors; all-horizon runs train for 280k. Evaluation uses exact-output pass@$k$ and the cumulative checkpoint pooling (standard test-time compute in \ref{tab:ttc-protocols}): predictions from all eight fixed dihedral views are inverted and voted at each 20k-step evaluation, giving 200 votes per held-out pair at 500k and 112 at 280k. We average the four held-out pairs within each task and then average across tasks; reported errors are sample standard deviations across retained replicas.

\paragraph{Held-out combinations.}
Table~\ref{tab:car-single-horizon} reports the terminal 500k evaluation. Independent full and low-rank tables fit the trained combinations but cannot transfer to a new rule--horizon key. Both structured memories transfer nontrivially. \method{} gives the strongest seen accuracy, while composition alone is better on unseen combinations, consistent with the private residual encouraging specialization to the trained horizon. For every unseen combination, the residual lookup receives a zero row, so \method{} uses only its compositional path at inference; it can nevertheless differ from the composition-only model because nonzero residuals can enter whenever the model chooses to use them.

\begin{table}[ht]
  \centering
  \caption{\textbf{CAR single-horizon generalization at 500k steps.} Models train on one horizon for each of 1{,}500 rules and evaluate both those seen combinations and the other 4{,}500 rule--horizon combinations. Parameters include the 1.710M shared backbone.}
  \label{tab:car-single-horizon}
  \small
  \resizebox{\linewidth}{!}{%
  \smalluncertainties
  \begin{tabular}{lrrcccc}
    \toprule
    & & & \multicolumn{2}{c}{Seen} & \multicolumn{2}{c}{Unseen}\\
    \cmidrule(lr){4-5}\cmidrule(lr){6-7}
    Task memory & Total params & Memory params & pass@1 & pass@2 & pass@1 & pass@2\\
    \midrule
    Full table & 4.783M & 3.072M & $50.78\pm1.56$ & $53.94\pm1.62$ & $0.00\pm0.00$ & $0.00\pm0.00$\\
    Low-rank table & 1.906M & 0.196M & $51.43\pm1.81$ & $53.49\pm1.76$ & $0.05\pm0.12$ & $0.07\pm0.19$\\
    Three-factor composition & 2.229M & 0.519M & $56.48\pm2.16$ & $57.89\pm2.03$ & $\mathbf{2.54}\pm0.52$ & $\mathbf{3.69}\pm0.56$\\
    Three-factor \method{} & 2.425M & 0.715M & $\mathbf{57.40}\pm0.97$ & $\mathbf{58.86}\pm0.90$ & $1.80\pm0.37$ & $2.73\pm0.48$\\
    \bottomrule
  \end{tabular}%
  }
\end{table}

\FloatBarrier

\begin{figure}[ht]
  \centering
  \includegraphics[width=0.78\linewidth]{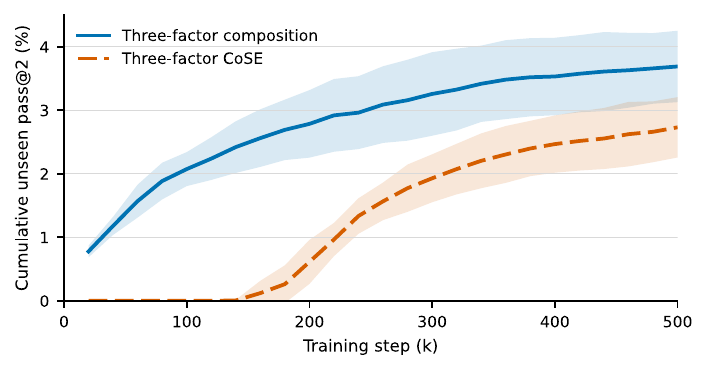}
  \caption{\textbf{Unseen generalization emerges later when composition is combined with a per-instance residual table.} Curves show the mean cumulative unseen pass@2 across seven runs; shaded bands indicate $\pm$ one sample standard deviation.}
  \label{fig:car-late-emergence}
\end{figure}

\paragraph{Late-emerging generalization.}
Figure~\ref{fig:car-late-emergence} shows the behavior that motivated the longer single-horizon training run. The three-factor composition-only setup generalizes from the beginning, and its mean cumulative unseen pass@2 exceeds 1\% by 40k steps. In contrast, \method{} remains at 0\% through 120k steps and first exceeds 1\% at 240k steps. This delayed emergence supports the hypothesis that, when composition and an instance-specific residual are both available, the network faces less pressure to learn a transferable factorization early: it initially favors specialization to observed combinations and only later learns a reusable factorization.

\FloatBarrier

Table~\ref{tab:car-unseen-horizon} breaks down the two transferable setups by the \emph{evaluation} horizon. For example, $T_{\rm eval}=1$ contains rules whose trained horizon was 2, 4, or 8. Composition wins at all four horizons; its aggregate pass@2 advantage over \method{} is 0.96 points.

\begin{table}[ht]
  \centering
  \caption{\textbf{CAR transfer by unseen evaluation horizon.} Results are over held-out rule--horizon combinations; both experiments use 7 repeats.}
  \label{tab:car-unseen-horizon}
  \small
  \smalluncertainties
  \begin{tabular}{lcccc}
    \toprule
    & \multicolumn{2}{c}{Three-factor composition} & \multicolumn{2}{c}{Three-factor \method{}}\\
    \cmidrule(lr){2-3}\cmidrule(lr){4-5}
    $T_{\rm eval}$ & pass@1 & pass@2 & pass@1 & pass@2\\
    \midrule
    1 & $\mathbf{1.78}\pm0.75$ & $\mathbf{2.87}\pm1.06$ & $1.00\pm0.37$ & $1.61\pm0.52$\\
    2 & $\mathbf{1.92}\pm0.83$ & $\mathbf{2.95}\pm0.93$ & $1.44\pm0.46$ & $2.20\pm0.37$\\
    4 & $\mathbf{3.49}\pm0.64$ & $\mathbf{4.79}\pm0.54$ & $2.43\pm0.50$ & $3.71\pm0.76$\\
    8 & $\mathbf{2.96}\pm0.57$ & $\mathbf{4.15}\pm0.40$ & $2.35\pm0.43$ & $3.41\pm0.56$\\
    \bottomrule
  \end{tabular}
\end{table}

\FloatBarrier

\paragraph{All-horizons control.}
When every rule--horizon combination is observed, three-factor \method{} has the highest mean accuracy (Table~\ref{tab:car-all-horizons}) and uses 1.291M memory parameters, compared with 12.288M for the full table. Three-factor composition uses the same 0.519M memory on the 1.5k and 6k datasets, demonstrating how shared factors avoid linear growth in the number of combinations.

\begin{table}[ht]
  \centering
  \caption{\textbf{Matched-budget CAR comparison at 280k steps.} The single-horizon dataset has 1{,}500 combinations; the all-horizons dataset has 6{,}000.}
  \label{tab:car-all-horizons}
  \footnotesize
  \setlength{\tabcolsep}{3.5pt}
  \resizebox{\linewidth}{!}{%
  \smalluncertainties
  \begin{tabular}{lrrrrrr}
    \toprule
    & \multicolumn{2}{c}{Memory params} & \multicolumn{2}{c}{Single-horizon} & \multicolumn{2}{c}{All-horizons}\\
    \cmidrule(lr){2-3}\cmidrule(lr){4-5}\cmidrule(lr){6-7}
    Task memory & 1.5k & 6k & pass@1 & pass@2 & pass@1 & pass@2\\
    \midrule
    Full table & 3.072M & 12.288M & $45.20\pm1.78$ & $48.24\pm2.07$ & $43.66\pm1.32$ & $47.21\pm1.72$\\
    Low-rank table & 0.196M & 0.772M & $45.09\pm1.46$ & $47.19\pm1.52$ & $43.09\pm0.31$ & $45.61\pm0.33$\\
    Two-factor composition & 0.518M & 1.670M & $52.41\pm1.65$ & $53.98\pm1.56$ & $48.20\pm1.04$ & $50.13\pm1.06$\\
    Two-factor \method{} & 0.714M & 2.442M & $51.97\pm2.36$ & $53.56\pm2.21$ & $47.53\pm1.81$ & $49.38\pm1.81$\\
    Three-factor composition & 0.519M & 0.519M & $50.26\pm2.38$ & $51.83\pm2.33$ & $49.43\pm1.82$ & $51.39\pm2.02$\\
    Three-factor \method{} & 0.715M & 1.291M & $51.80\pm1.96$ & $53.37\pm1.99$ & $\mathbf{50.45}\pm1.28$ & $\mathbf{52.25}\pm1.23$\\
    \bottomrule
  \end{tabular}
  }
\end{table}

\paragraph{Comparison with frontier LLM.}
Table~\ref{tab:car-gpt} compares the 500k three-factor \method{} model against GPT-5.6 (\texttt{gpt-5.6-sol}, reasoning effort xhigh) on the seen combinations, grouped by horizon. GPT-5.6 was evaluated on 25 puzzles per horizon with two independent attempts each, with the demonstration pairs and all test inputs inlined in a single prompt, either without tools or with a sandboxed Python interpreter. Without code execution, GPT-5.6 degrades sharply beyond one step, while the CAR model retains substantial accuracy through $T{=}4$; with Python, GPT-5.6 can simulate candidate rules and overtakes the from-scratch model at long horizons. 

\begin{table}[ht]
  \centering
  \caption{\textbf{CAR accuracy by rollout horizon: three-factor \method{} vs.\ GPT-5.6.} CAR values are means over seven replicas of the terminal cumulative evaluation on seen combinations; GPT-5.6 entries have no across-run error bars.}
  \label{tab:car-gpt}
  \small
  \resizebox{\linewidth}{!}{%
  \begin{tabular}{lrrrrrrrr}
    \toprule
    & \multicolumn{2}{c}{$T{=}1$} & \multicolumn{2}{c}{$T{=}2$} & \multicolumn{2}{c}{$T{=}4$} & \multicolumn{2}{c}{$T{=}8$}\\
    \cmidrule(lr){2-3}\cmidrule(lr){4-5}\cmidrule(lr){6-7}\cmidrule(lr){8-9}
    Method & pass@1 & pass@2 & pass@1 & pass@2 & pass@1 & pass@2 & pass@1 & pass@2\\
    \midrule
    CAR, three-factor \method{} & $\mathbf{98.70}$ & $\mathbf{99.25}$ & $\mathbf{88.84}$ & $\mathbf{90.27}$ & 31.56 & 34.15 & 10.50 & 11.77\\
    GPT-5.6, no tools & 48.0 & 66.0 & 8.5 & 11.0 & 3.0 & 5.0 & 5.0 & 8.0\\
    GPT-5.6, Python & 78.5 & 90.0 & 73.5 & 78.0 & $\mathbf{66.5}$ & $\mathbf{72.0}$ & $\mathbf{66.0}$ & $\mathbf{76.0}$\\
    \bottomrule
  \end{tabular}%
  }
\end{table}

\FloatBarrier

\subsection{Language modeling with compositional n-gram memory}
\label{app:lm}

\paragraph{Architecture.}
We implement Engram~\citep{cheng2026engram} in nanochat~\citep{karpathy2025nanochat}. Engram canonicalizes token identities, hashes 2- and 3-gram contexts through eight heads into learned lookup tables, and passes the retrieved features through its standard contextual gate and short convolution. Our \method{} variant retains this path and adds a shared FiLM branch that composes the embeddings of the current token and the two preceding tokens, its output is concatenated with the retrieved features. These embeddings are detached views of the trunk token table, so the composer adds no second vocabulary table and cannot perturb the embedding through this branch. The comparison therefore changes how the same structured n-gram key is represented without introducing explicit task identities.

More precisely, let $\mathbf e_t\in\mathbb R^D$ be the detached embedding of token $x_t$, and let $\mathbf m_t^{(n)}\in\mathbb R^w$ denote the concatenated 8-head lookup for its canonicalized $n$-gram context. The language-model \method{} composer and the resulting Engram feature are
\begin{align}
  \mathbf r_t &= W_r\mathbf e_t,
  & \mathbf c_t &= W_c[\mathbf e_{t-1};\mathbf e_{t-2}], \nonumber\\
  \mathbf s_t &= \operatorname{RMSNorm}\!\left((\mathbf 1+W_\gamma\mathbf c_t)\odot\mathbf r_t+W_\beta\mathbf c_t+\mathbf c_t\right),
  & \mathbf h_t^{\textsc{CoSE}} &= [\mathbf s_t;\mathbf m_t^{(2)};\mathbf m_t^{(3)}].
  \label{eq:lm-cose}
\end{align}
Here $[\,;\,]$ denotes concatenation and $\mathbf s_t\in\mathbb R^p$. \method{}-Mixture is a variant that projects each lookup into the composed space and mixes the results using a router conditioned on $\mathbf s_t$:
\begin{align}
  \boldsymbol\alpha_t
    &= \operatorname{softmax}(W_a\mathbf s_t), \nonumber\\
  \mathbf h_t^{\mathrm{mix}}
    &= \operatorname{RMSNorm}\!\left(
      \mathbf s_t+\sum_{n\in\{2,3\}}\alpha_{t,n}U_n\mathbf m_t^{(n)}
    \right),
  \label{eq:lm-cose-mixture}
\end{align}
where $U_n\in\mathbb R^{p\times w}$, $\boldsymbol\alpha_t\in\mathbb R^2$, and all $W$ and $U_n$ are learned linear maps. In both variants, $\mathbf h_t$ is passed to Engram's unchanged key--value gate and short convolution.

\paragraph{Experiments.}
The primary sweep uses a depth-12, approximately 286M-parameter backbone, a 524{,}288-token global batch, and a target parameter-to-data ratio of 10.5 ($\sim$1.19B tokens). Evaluation uses 10{,}485{,}760 validation tokens every 250 steps. We use the nanochat default optimizer schedule with a $5\times$ learning-rate multiplier on memory parameters. Every experiment is repeated with five seeds; data order is fixed across seeds, and the primary metric is final validation bits per byte.

We study the scaling of n-gram memory along two axes: the number of lookup rows and the width of each row. Table~\ref{tab:nanochat-d12} varies the row-capacity multiplier at the full width of 384, while Table~\ref{tab:nanochat-d12-width} varies width at a fixed row-capacity multiplier of $\times5$. The row-capacity multiplier $\times k$ sets each lookup table's row count to a distinct prime just above $k|\mathcal V|$, where $|\mathcal V|=32{,}768$ is the tokenizer vocabulary size.

\begin{table}[ht]
  \centering
  \caption{\textbf{\method{} improves n-gram memory across row capacities on nanochat d12.} All nonzero rows use five seeds.}
  \label{tab:nanochat-d12}
  \small
  \smalluncertainties
  \begin{tabular}{lrrrr}
    \toprule
    & \multicolumn{2}{c}{Memory params} & \multicolumn{2}{c}{Validation bpb}\\
    \cmidrule(lr){2-3}\cmidrule(lr){4-5}
    Capacity & Engram & \method{} & Engram & \method{}\\
    \midrule
    0 & 0 & 3.5M & $0.85365\pm0.00111$ & $0.85363\pm0.00051$\\
    $\times5$ & 254M & 258M & $0.84638\pm0.00050$ & $\mathbf{0.84436}\pm0.00083$\\
    $\times10$ & 506M & 510M & $0.84482\pm0.00057$ & $\mathbf{0.84277}\pm0.00081$\\
    $\times20$ & 1.01B & 1.01B & $0.84333\pm0.00046$ & $\mathbf{0.84103}\pm0.00064$\\
    $\times40$ & 2.02B & 2.02B & $0.84222\pm0.00062$ & $\mathbf{0.83946}\pm0.00058$\\
    \bottomrule
  \end{tabular}
\end{table}

At zero table capacity, Engram reduces to the stock backbone, whereas \method{} becomes composition only. Their nearly identical results show that composition alone brings almost no advantage and instead helps through its interaction with the memory table. At every nonzero row capacity, \method{} has lower mean bpb and wins all 20 paired-seed comparisons, and it matches or exceeds Engram with twice the row capacity.

\begin{table}[ht]
  \centering
  \caption{\textbf{\method{}-Mixture performs best across lookup widths on nanochat d12.} The row-capacity multiplier is fixed at $\times5$; all cells use five seeds. Width is the lookup dimension per n-gram order.}
  \label{tab:nanochat-d12-width}
  \small
  \setlength{\tabcolsep}{2.8pt}
  \smalluncertainties
  \begin{tabular}{lrrrrrr}
    \toprule
    & \multicolumn{3}{c}{Memory params} & \multicolumn{3}{c}{Validation bpb}\\
    \cmidrule(lr){2-4}\cmidrule(lr){5-7}
    Width & Engram & \method{} & \method{}-Mix & Engram & \method{} & \method{}-Mix\\
    \midrule
    24 & 15.9M & 19.4M & 19.3M & $0.85067\pm0.00097$ & $0.84937\pm0.00090$ & $\mathbf{0.84854}\pm0.00042$\\
    48 & 31.8M & 35.3M & 35.1M & $0.84878\pm0.00099$ & $0.84773\pm0.00108$ & $\mathbf{0.84695}\pm0.00047$\\
    96 & 63.6M & 67.1M & 66.7M & $0.84685\pm0.00063$ & $0.84569\pm0.00109$ & $\mathbf{0.84504}\pm0.00032$\\
    \bottomrule
  \end{tabular}
\end{table}

\FloatBarrier

\begin{figure}[ht]
  \centering
  \includegraphics[width=\linewidth]{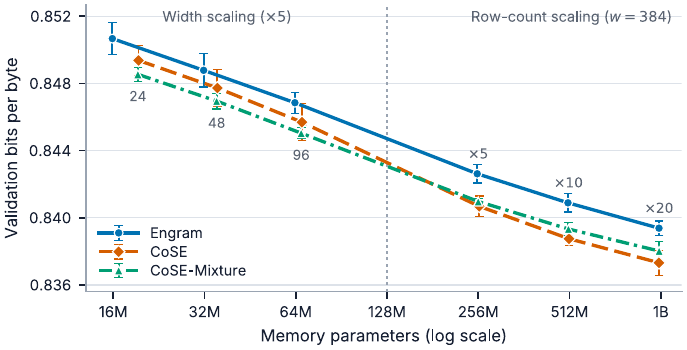}
  \caption{\textbf{\method{} and \method{}-Mixture consistently outperform Engram across memory scales on nanochat d12.} The width- and row-count-scaling ablations share a log-scaled parameter axis, each point is the mean across five seeds.}
  \label{fig:lm-memory-scaling}
\end{figure}

Figure~\ref{fig:lm-memory-scaling} summarizes both d12 ablations on a memory-parameter axis. Both compositional variants achieve lower mean bpb than Engram at every width. \method{}-Mixture achieves the lowest mean throughout and matches or exceeds the performance of Engram with twice the row width. However, \method{}-Mixture performs worse than \method{} when scaling the number of rows. Therefore we do not propose it as a superior replacement for \method{} in language modeling. The gains from scaling row width are smaller than those observed in the row-capacity study, suggesting that the parameter-efficiency benefit is greater when memory is scaled by increasing the number of rows rather than the row width.

\FloatBarrier

\paragraph{Backbone scaling and token-budget controls.}
We repeat the row-capacity study at the full width of 640 with a depth-20, 896.5M-parameter backbone, using the same 32k-token tokenizer and validation shard as d12. The global batch is 1{,}048{,}576 tokens, and the compute-optimal training budget is roughly 4.64B tokens. Evaluation and memory learning-rate settings match d12. \method{} remains ahead at all three capacities, although the margin is smaller and does not grow monotonically; $\times40$ exceeds GPU VRAM even at device batch one and is therefore omitted. The corresponding fixed-$\times5$ width study uses widths 40, 80, and 160. \method{} still has lower mean bpb but smaller matched-width advantages than d12.

\begin{table}[ht]
  \centering
  \caption{\textbf{Matched-capacity language-model results at d20.} All experiments use five seeds.}
  \label{tab:nanochat-d20}
  \small
  \smalluncertainties
  \begin{tabular}{lrrrr}
    \toprule
    & \multicolumn{2}{c}{Memory params} & \multicolumn{2}{c}{Validation bpb}\\
    \cmidrule(lr){2-3}\cmidrule(lr){4-5}
    Capacity & Engram & \method{} & Engram & \method{}\\
    \midrule
    $\times5$ & 426M & 436M & $0.73002\pm0.00031$ & $\mathbf{0.72939}\pm0.00082$\\
    $\times10$ & 846M & 856M & $0.72901\pm0.00045$ & $\mathbf{0.72800}\pm0.00091$\\
    $\times20$ & 1.68B & 1.69B & $0.72730\pm0.00019$ & $\mathbf{0.72664}\pm0.00058$\\
    \bottomrule
  \end{tabular}
\end{table}

\FloatBarrier

The d20 scale-up changes both backbone size and absolute token count. To isolate the effect of training data, the token-budget control evaluates each backbone with 1.19B and 4.64B tokens while holding its $\times5$ memory configuration fixed (Table~\ref{tab:nanochat-token-budget}). At 1.19B tokens, d12 and d20 have similar \method{}--Engram gaps; increasing the budget to 4.64B tokens shrinks the gap for both backbones. These small-$n$ controls suggest that the apparent scale dependence may largely reflect the training-data budget: \method{} has the largest advantage when data are scarce relative to the memory's structured key space. Conversely, expanding that key space, for example by increasing the number or diversity of n-gram patterns, may increase the relative advantage of \method{}.

\begin{table}[ht]
  \centering
  \caption{\textbf{Token-budget controls at fixed $\times5$ memory capacity.} Cells report paired \method{} minus Engram validation bpb; lower is better.}
  \label{tab:nanochat-token-budget}
  \small
  \smalluncertainties
  \begin{tabular}{lcc}
    \toprule
    Backbone & 1.19B tokens & 4.64B tokens\\
    \midrule
    d12 & $-0.0020\pm0.0005$ ($n=5$) & $-0.0002\pm0.0003$ ($n=2$)\\
    d20 & $-0.0018\pm0.0003$ ($n=3$) & $-0.0006\pm0.0008$ ($n=5$)\\
    \bottomrule
  \end{tabular}
\end{table}

The downstream CORE metric is noisier across seeds and does not consistently track the small bpb differences, so we focus on validation bpb. Together, the width and row-capacity scaling studies establish transfer beyond ARC's group transformations, while the token-budget control suggests that composition helps most when data are scarce relative to the hashed memory's structured key space.

\FloatBarrier

\section{Unseen-task adaptation and continual learning}
\label{app:cose-cl}

This appendix evaluates adaptation to unseen ARC-AGI-2 puzzles from ARC-AGI-1 base checkpoints under two complementary settings. In \emph{per-puzzle fine-tuning}, the base checkpoint is restored for each puzzle and the full model is adapted independently, measuring task-level adaptation without a retention constraint. In \emph{continual learning}, one model processes the same puzzles sequentially and must retain both earlier stream tasks and its original ARC-AGI-1 capability; we compare frozen private-row adaptation with naive fine-tuning, rehearsal, EWC, and L2-SP. Across both settings, optimization uses only each puzzle's demonstration pairs, while held-out outputs are reserved for evaluation.

\subsection{Protocol and baselines}

\paragraph{Base checkpoints and streams.}
The seven base models are 600k-step ARC-AGI-1 checkpoints pretrained with Muon. We use \method{} with residual ranks $r=0,32,512$ and the default 512-dimensional per-task embedding in the compositional branch ($t=512$), alongside the baseline full tables with $r=32,512$. Two additional \method{} variants reduce that embedding from 512 to 32 dimensions ($t=32$), at residual ranks $r=0$ and $r=32$ to push parameter efficiency to the limit. We use two streams: 233 ARC-AGI-2 training puzzles and 114 public-evaluation puzzles unseen to the base checkpoints. Table~\ref{tab:cose-cl-zero} reports their starting accuracies.

\method{} with $r=0$ is \emph{composition only}: it has no residual table, but still trains a small task embedding for each new puzzle. All augmentations of that puzzle share the same embedding. So its memory-only adaptation trains 512 parameters per puzzle, or 32 in the reduced-width variant. The latter uses a pretrained 32 to 512 projection that remains frozen during adaptation. For $r>0$, the current puzzle's new residual rows are also trained; shared composition parameters remain fixed. Table~\ref{tab:cose-cl-params} gives the resulting counts.

\begin{table}[htbp]
  \centering
  \caption{\textbf{Base checkpoints and zero-shot starting points.} ARC-1 uses the fixed 60-puzzle evaluation; stream columns report mean per-puzzle pass@1 and pass@2 before adaptation.}
  \label{tab:cose-cl-zero}
  \footnotesize
  \setlength{\tabcolsep}{3pt}
  \smalluncertainties
  \begin{tabular}{@{}lccccc@{}}
    \toprule
    & ARC-1 accuracy & \multicolumn{2}{c}{ARC-2 train zero-shot} & \multicolumn{2}{c}{ARC-2 eval zero-shot}\\
    \cmidrule(lr){3-4}\cmidrule(lr){5-6}
    Base task memory & pass@2 & pass@1 & pass@2 & pass@1 & pass@2\\
    \midrule
    \method{} $(r=0,t=32)$ & 77.50 & 0.86 & 0.86 & 0.00 & 0.00\\
    \method{} $(r=0)$ & 78.33 & 0.86 & 0.86 & 0.00 & 0.00\\
    \method{} $(r=32,t=32)$ & 77.50 & 0.43 & 0.43 & 0.00 & 0.00\\
    \method{} $(r=32)$ & 80.00 & 0.43 & 0.86 & 0.00 & 0.00\\
    \method{} $(r=512)$ & 76.67 & 0.43 & 0.86 & 0.00 & 0.00\\
    Table $(r=32)$ & 75.00 & 1.29 & 1.50 & 0.00 & 0.00\\
    Table $(r=512)$ & 78.33 & 0.86 & 1.72 & 0.00 & 0.00\\
    \bottomrule
  \end{tabular}
\end{table}

\begin{table}[htbp]
  \centering
  \caption{\textbf{Parameters updated by memory-only adaptation.} Each puzzle has $m$ augmentation instances, averaging 976.42 on the training stream and 996.28 on public evaluation. Shared weights are fixed. Composition-only \method{} ($r=0$) trains a task embedding but no residual rows.}
  \label{tab:cose-cl-params}
  \footnotesize
  \setlength{\tabcolsep}{3pt}
  \smalluncertainties
  \begin{tabular}{@{}lcccc@{}}
    \toprule
    Base task memory & Task embedding row & Residual rows & Train params/task & Eval params/task\\
    \midrule
    \method{} $(r=0,t=32)$ & 32 & --- & 32 & 32\\
    \method{} $(r=0)$ & 512 & --- & 512 & 512\\
    \method{} $(r=32,t=32)$ & 32 & $32m$ & 31.28k & 31.91k\\
    \method{} $(r=32)$ & 512 & $32m$ & 31.76k & 32.39k\\
    \method{} $(r=512)$ & 512 & $512m$ & 500.44k & 510.61k\\
    Table $(r=32)$ & --- & $32m$ & 31.25k & 31.88k\\
    Table $(r=512)$ & --- & $512m$ & 499.93k & 510.10k\\
    \bottomrule
  \end{tabular}
\end{table}

\paragraph{Optimization and evaluation.}
Each puzzle receives 1{,}000 updates with global batch 128 and a 20-step warmup. Fresh task embedding rows use row-local AdamW at learning rate $10^{-2}$ with no weight decay; instance rows use row-local SignSGD at $10^{-2}$ with weight decay 0.1. We have also experimented with optimizing the residual instance rows with AdamW but found it to be less effective. When a method updates shared weights, it uses AdamW at $10^{-4}$ with weight decay 0.1 and $\beta=(0.9,0.95)$. Muon is used for base-model pretraining. We use auxiliary halting-loss weight $w_h=0$ while retaining adaptive training halting. Training samples use the standard ARC-AGI augmentations in this paper. Evaluation takes the first 128 offline augmentations, removes each augmentation from the predicted grid, and ranks unique predictions by votes and confidence. We average exact pass@$k$ across held-out outputs within each puzzle and then across puzzles. Full-stream rechecks occur every 15 stages and at the end; the ARC-AGI-1 stability test uses 64 augmentations.

\paragraph{Methods.}
\emph{Memory-only CL} retains newly trained puzzle rows while keeping the backbone, shared composition parameters, base-model rows, and all previously learned rows fixed. \emph{Per-puzzle FT Oracle} restores the base checkpoint for each puzzle and fine-tunes the full model independently; it has no single retained model. \emph{Joint fine-tuning} trains one full model on demonstrations from all stream puzzles together for $1{,}000N$ steps. Naive fine-tuning, EWC~\citep{kirkpatrick2017ewc}, L2-SP~\citep{xuhong2018l2sp}, and rehearsal, update either the composition module in \method{} or all shared weights, always together with the current private rows. These baselines use the \method{} $(r=32)$ checkpoint on the training stream. For EWC we estimate a diagonal Fisher from 64 ARC-AGI-1 pretraining batches and use $\lambda=10^{-2}$; L2-SP anchors weights to the base checkpoint with $\lambda=10^{-1}$. Rehearsal replaces every second update with an ARC-AGI-1 pretraining batch, keeping the number of updates fixed.

\FloatBarrier
\subsection{Per-task adaptation and parameter efficiency}

\method{} substantially improves adaptation to unseen tasks through new memory rows while keeping shared weights fixed (Table~\ref{tab:cose-cl-core}). Memory-only \method{} $(r=32)$ reaches $43.58\pm0.58$\% pass@2 on training puzzles and $12.19\pm0.90$\% on public-evaluation puzzles, compared with 19.73\% and 4.47\% for the full table $(r=512)$. This gives $2.2\times$ and $2.7\times$ the full table's accuracy while updating roughly $1/16$ as many parameters per puzzle. Yet under the Per-puzzle FT Oracle, the full table reaches 49.42\% and 15.94\%, comparable to the 46.27\% and 15.20\% of \method{} $(r=32)$. The full-table model therefore has sufficient capacity to reach comparable accuracy; it is not able to adapt efficiently with new memory rows.

\begin{figure}[htbp]
  \centering
  \includegraphics[width=0.78\linewidth]{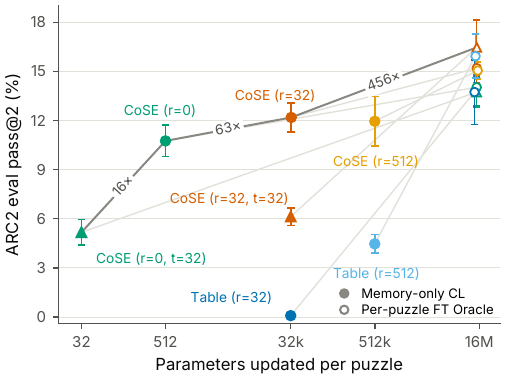}
  \caption{\textbf{Parameter efficiency on unseen ARC-AGI-2 public-evaluation puzzles.} Composition-only \method{} ($r=0$) provides strong adaptation with 32 or 512 parameters per puzzle; the rank-32 residual raises accuracy with the default task embedding width (512). Filled markers denote memory-only CL and open markers the Per-puzzle FT Oracle. The line connects four representative Pareto points, with labels giving parameter ratios between adjacent points. Error bars show sample standard deviations.}
  \label{fig:cl-param-efficiency-eval}
\end{figure}

We can push this parameter efficiency much further. Most of this adaptation performance can be retained with just a single 512-parameter task embedding per puzzle. Composition-only \method{} $(r=0)$ reaches $39.73\pm1.48$\% and $10.76\pm0.96$\% pass@2, retaining 91\% and 88\% of the accuracy of memory-only \method{} $(r=32)$ on the two streams. It still more than doubles the full table's performance while updating roughly $1/1{,}000$ as many parameters. Thus, the compositional task embedding provides the main mechanism for efficient adaptation, with residual rows contributing an additional accuracy gain. The embedding-width ablation reinforces this interpretation: reducing the task embedding of \method{} $(r=32)$ to 32 dimensions saves only 480 parameters per puzzle but sharply reduces accuracy despite retaining the residual rows. Figures~\ref{fig:cose-cl}(a) and~\ref{fig:cl-param-efficiency-eval} show the resulting accuracy--parameter trade-offs. These gains also preserve prior knowledge: all 140 memory-only runs retain their original ARC-AGI-1 accuracy with zero measured drift in shared parameters and previously learned rows.

\begin{table}[htbp]
  \centering
  \caption{\textbf{Held-out-task adaptation from ARC-AGI-1 base checkpoints.} All rows use $w_h=0$. $r=0$ denotes composition only, and $t=512$ unless specified. Counts are effective updated parameters per puzzle.}
  \label{tab:cose-cl-core}
  \footnotesize
  \setlength{\tabcolsep}{2.5pt}
  \smalluncertainties
  \begin{tabular}{@{}lrccrcc@{}}
    \toprule
    & \multicolumn{3}{c}{Memory-only CL} & \multicolumn{3}{c}{Per-puzzle FT Oracle}\\
    \cmidrule(lr){2-4}\cmidrule(l){5-7}
    Base task memory & Params/task & pass@1 & pass@2 & Params/task & pass@1 & pass@2\\
    \midrule
    \multicolumn{7}{@{}l}{\emph{ARC-AGI-2 training stream}}\\
    \method{} $(r=0,t=32)$ & 32 & $16.78\pm0.44$ & $20.30\pm0.44$ & 14.733M & $42.93\pm1.22$ & $48.70\pm1.85$\\
    \method{} $(r=0)$ & 512 & $36.05\pm1.15$ & $39.73\pm1.48$ & 14.717M & $42.23\pm1.28$ & $47.02\pm2.15$\\
    \method{} $(r=32,t=32)$ & 31.28k & $19.40\pm0.89$ & $23.76\pm0.26$ & 14.781M & $\mathbf{45.23}\pm0.83$ & $49.25\pm1.20$\\
    \method{} $(r=32)$ & 31.76k & $\mathbf{38.56}\pm0.70$ & $\mathbf{43.58}\pm0.58$ & 14.765M & $41.93\pm0.44$ & $46.27\pm1.32$\\
    \method{} $(r=512)$ & 500.44k & $37.81\pm0.77$ & $42.26\pm1.31$ & 15.217M & $43.35\pm1.59$ & $48.51\pm2.02$\\
    Table $(r=32)$ & 31.25k & $5.19\pm0.24$ & $7.77\pm0.24$ & 13.710M & $40.99\pm3.33$ & $46.49\pm0.45$\\
    Table $(r=512)$ & 499.93k & $15.41\pm0.78$ & $19.73\pm0.49$ & 14.163M & $43.40\pm1.02$ & $\mathbf{49.42}\pm1.80$\\
    \midrule
    \multicolumn{7}{@{}l}{\emph{ARC-AGI-2 public-evaluation stream}}\\
    \method{} $(r=0,t=32)$ & 32 & $3.86\pm0.84$ & $5.18\pm0.78$ & 14.733M & $10.58\pm1.18$ & $13.74\pm0.91$\\
    \method{} $(r=0)$ & 512 & $9.21\pm1.12$ & $10.76\pm0.96$ & 14.717M & $12.33\pm0.61$ & $14.04\pm1.14$\\
    \method{} $(r=32,t=32)$ & 31.91k & $5.88\pm0.59$ & $6.14\pm0.54$ & 14.782M & $13.74\pm1.01$ & $\mathbf{16.47}\pm1.69$\\
    \method{} $(r=32)$ & 32.39k & $11.05\pm1.56$ & $\mathbf{12.19}\pm0.90$ & 14.766M & $13.40\pm0.97$ & $15.20\pm0.88$\\
    \method{} $(r=512)$ & 510.61k & $\mathbf{11.46}\pm1.08$ & $11.96\pm1.52$ & 15.228M & $13.16\pm0.25$ & $15.06\pm0.51$\\
    Table $(r=32)$ & 31.88k & $0.00\pm0.00$ & $0.09\pm0.20$ & 13.711M & $10.77\pm2.13$ & $13.74\pm1.98$\\
    Table $(r=512)$ & 510.10k & $3.60\pm0.57$ & $4.47\pm0.57$ & 14.173M & $\mathbf{13.89}\pm0.76$ & $15.94\pm1.34$\\
    \bottomrule
  \end{tabular}
\end{table}

\paragraph{Joint fine-tuning.}
Table~\ref{tab:cose-cl-joint} reports the offline single-model control. Without auxiliary halting loss, joint \method{} $(r=32)$ reaches $33.76\pm2.86$\% and $8.63\pm1.16$\%, below memory-only adaptation on both streams. Parameter counts amortize one set of shared weights and all puzzle-private rows across the stream. Composition-only $t=512$ reaches $31.35\pm2.60$\% on training puzzles and $7.60\pm0.83$\% on public-evaluation puzzles; joint \method{} $(r=512)$ reaches $31.59\pm4.89$\% on training puzzles.

\begin{table}[htbp]
  \centering
  \caption{\textbf{Joint fine-tuning has simultaneous access to all stream puzzles.} All rows use $w_h=0$. Parameter counts include shared weights and all puzzle-private rows, divided by the number of puzzles.}
  \label{tab:cose-cl-joint}
  \footnotesize
  \setlength{\tabcolsep}{3pt}
  \smalluncertainties
  \begin{tabular}{@{}lrccc@{}}
    \toprule
    Base task memory & Params/task & $w_h$ & pass@1 & pass@2\\
    \midrule
    \multicolumn{5}{@{}l}{\emph{ARC-AGI-2 training stream}}\\
    \method{} $(r=0,t=32)$ & 63.27k & 0 & $26.02\pm2.00$ & $31.88\pm0.58$\\
    \method{} $(r=0)$ & 63.67k & 0 & $26.31\pm1.85$ & $31.35\pm2.60$\\
    \method{} $(r=32,t=32)$ & 94.58k & 0 & $29.08\pm0.23$ & $32.83\pm0.45$\\
    \method{} $(r=32)$ & 94.99k & 0 & $28.06\pm3.98$ & $33.76\pm2.86$\\
    \method{} $(r=512)$ & 563.60k & 0 & $25.83\pm3.43$ & $31.59\pm4.89$\\
    Table $(r=32)$ & 89.95k & 0 & $19.10\pm1.50$ & $23.62\pm0.59$\\
    Table $(r=512)$ & 558.57k & 0 & $18.41\pm3.05$ & $21.49\pm2.79$\\
    \midrule
    \multicolumn{5}{@{}l}{\emph{ARC-AGI-2 public-evaluation stream}}\\
    \method{} $(r=0,t=32)$ & 129.27k & 0 & $6.09\pm1.14$ & $6.77\pm1.85$\\
    \method{} $(r=0)$ & 129.61k & 0 & $6.07\pm0.10$ & $7.60\pm0.83$\\
    \method{} $(r=32,t=32)$ & 161.30k & 0 & $5.95\pm1.04$ & $8.63\pm0.15$\\
    \method{} $(r=32)$ & 161.63k & 0 & $5.75\pm0.94$ & $8.63\pm1.16$\\
    \method{} $(r=512)$ & 639.70k & 0 & $5.31\pm1.49$ & $7.80\pm1.39$\\
    Table $(r=32)$ & 151.87k & 0 & $4.24\pm0.25$ & $4.82\pm0.44$\\
    Table $(r=512)$ & 629.94k & 0 & $4.63\pm0.55$ & $5.17\pm1.48$\\
    \bottomrule
  \end{tabular}
\end{table}

\FloatBarrier
\subsection{Continual-learning comparison}

Table~\ref{tab:cose-cl-baselines} compares retained models on the same 233-puzzle stream, matching Figure~\ref{fig:cose-cl}(b). \emph{Immediate} accuracy is measured after each puzzle's own stage; \emph{final} accuracy evaluates the retained model at its reported endpoint. Memory-only \method{} reaches $43.37\pm0.48$\% final pass@2 and preserves the original 80.00\% ARC-AGI-1 accuracy. We next consider continual-learning baselines that update previously trained shared parameters alongside the new puzzle-specific rows. Without regularization, updating the shared composition module causes substantial forgetting: pass@2 falls from $40.72\pm2.12$\% immediately after adaptation to $6.22\pm2.07$\% at the endpoint, while ARC-AGI-1 accuracy falls to zero.

\begin{figure}[htbp]
  \centering
  \includegraphics[width=0.78\linewidth]{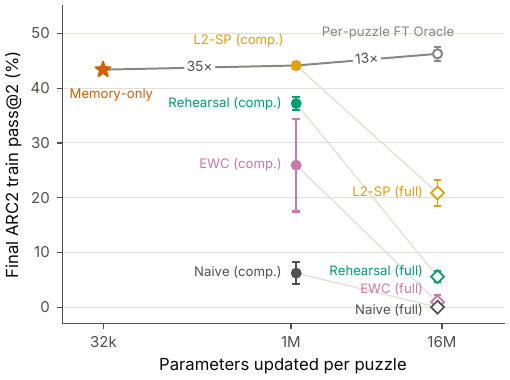}
  \caption{\textbf{Parameter efficiency of continual-learning methods on the same \method{} $(r=32)$ checkpoint.} Memory-only adaptation uses 31.8k parameters per puzzle, versus 1.10M for composition-scoped L2-SP and 14.77M for full-model updates. Light lines connect the two scopes of each baseline; the darker line connects the parameter--accuracy frontier, including the independent Per-puzzle FT Oracle. Ratios indicate relative parameter costs. Error bars show sample standard deviations.}
  \label{fig:cl-baseline-param-efficiency}
\end{figure}

Regularizing composition updates with L2-SP raises final pass@2 to $44.13\pm0.70$\%, 0.76 percentage points above memory-only adaptation. This small advantage over memory-only \method{} requires approximately $35\times$ as many updated parameters per puzzle (Figure~\ref{fig:cl-baseline-param-efficiency}). Extending L2-SP to the full model preserves a similar ARC-AGI-1 accuracy (77.78\%), but ARC-AGI-2 pass@2 drops from $52.37\pm2.38$\% immediately after adaptation to $20.86\pm2.34$\% at the endpoint. Preserving the pretrained capability alone therefore does not ensure retention of newly learned tasks.

\begin{table}[htbp]
  \centering
  \caption{\textbf{Stability--plasticity comparison on the ARC-AGI-2 training stream.} All methods use the \method{} $(r=32)$ checkpoint with $w_h=0$. }
  \label{tab:cose-cl-baselines}
  \footnotesize
  \setlength{\tabcolsep}{2pt}
  \smalluncertainties
  \begin{tabular}{@{}llrcccc@{}}
    \toprule
    Method & Scope & Params/task & Immediate & Final p@1 & Final p@2 & ARC-1 after\\
    \midrule
    Per-puzzle FT Oracle & full, per-task & 14.765M & $46.27\pm1.32$ & $41.93\pm0.44$ & $46.27\pm1.32$ & ---\\
    Memory-only & private rows & 31.76k & $42.94\pm1.47$ & $38.65\pm0.98$ & $43.37\pm0.48$ & $80.00\pm0.00$\\
    L2-SP & composition & 1.102M & $45.40\pm1.68$ & $38.87\pm0.86$ & $44.13\pm0.70$ & $78.33\pm0.00$\\
    Rehearsal & composition & 1.102M & $41.94\pm1.25$ & $31.76\pm1.67$ & $37.20\pm1.25$ & $78.06\pm3.76$\\
    EWC & composition & 1.102M & $41.99\pm1.99$ & $21.22\pm8.31$ & $25.94\pm8.46$ & $30.28\pm37.33$\\
    Naive & composition & 1.102M & $40.72\pm2.12$ & $4.79\pm1.43$ & $6.22\pm2.07$ & $0.00\pm0.00$\\
    L2-SP & full model & 14.765M & $52.37\pm2.38$ & $16.50\pm1.33$ & $20.86\pm2.34$ & $77.78\pm0.96$\\
    Rehearsal & full model & 14.765M & $45.51\pm0.98$ & $4.58\pm1.18$ & $5.58\pm0.94$ & $56.94\pm4.28$\\
    EWC & full model & 14.765M & $44.11\pm0.62$ & $0.79\pm1.36$ & $0.93\pm1.26$ & $6.67\pm11.55$\\
    Naive & full model & 14.765M & $5.22$ & $0.00\pm0.00$ & $0.00\pm0.00$ & $0.00$\\
    \bottomrule
  \end{tabular}
\end{table}

\begin{figure}[htbp]
  \centering
  \includegraphics[width=\linewidth]{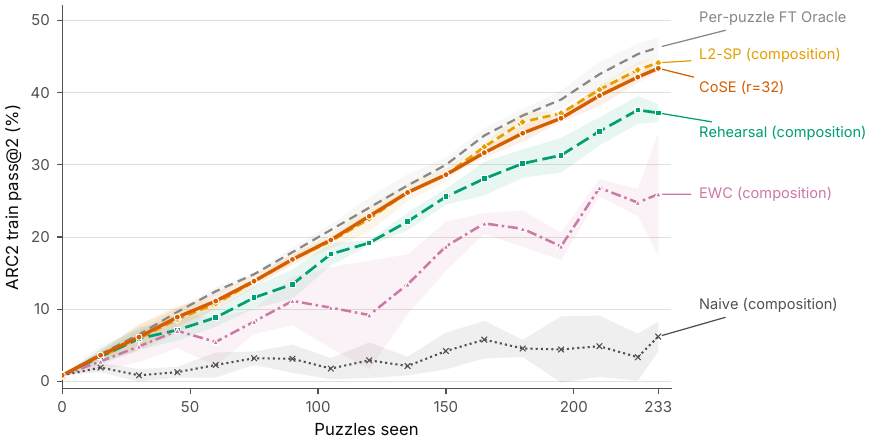}
  \caption{\textbf{Retained performance across the ARC-AGI-2 training stream.} We evaluate on the full stream after every 15 puzzles and at the endpoint. Shading shows sample standard deviation. Baselines update shared composition parameters, while memory-only \method{} updates only new private rows. The Per-puzzle FT Oracle aggregates independent adaptations.}
  \label{fig:cl-stream-dynamics}
\end{figure}

Figure~\ref{fig:cl-stream-dynamics} shows retained performance across the stream. Memory-only \method{} improves steadily as new puzzle memories are added. Composition-scoped L2-SP tracks it closely, while rehearsal, EWC, and naive composition updates retain less of the acquired performance.

\paragraph{Discussion.}
\method{} encourages a clearer separation between shared reasoning in the backbone and puzzle-specific knowledge in memory, supporting generalization to new tasks. We illustrate this with exploratory memory-only adaptation experiments. With shared weights frozen, we zeroed selected task memories and refitted them on subsets of previously learned ARC1 and new ARC2 puzzles. The full-table baseline reached 100\% accuracy on old puzzles versus 40.6\% on new puzzles; composition-only \method{} showed a smaller gap, at 87.5\% versus 75.0\%. This contrast suggests that the full-table baseline relies more on puzzle-specific knowledge embedded in its backbone, limiting adaptation to new tasks. Separately, adding a single trainable puzzle vector shared across augmented memory entries substantially improved the full-table baseline. This minimal form of \method{} provides the structural support needed to consolidate information across a puzzle's many memory entries.

\FloatBarrier
\subsection{Halting-loss ablation and complete results}
\label{app:cose-cl-halt}

We compare auxiliary halting-loss weights $w_h\in\{0,0.5\}$, keeping adaptive training halting enabled in both settings. Removing the auxiliary loss generally improves per-puzzle full-model adaptation, with lower token losses consistent with better fitting (Table~\ref{tab:cose-cl-token-loss}). Its effect on memory-only adaptation depends on the architecture, while joint fine-tuning usually benefits from retaining the loss. Tables~\ref{tab:cose-cl-halt-train-frozen}--\ref{tab:cose-cl-halt-full} report complete results for both settings across architectures and adaptation protocols.

\begin{table}[htbp]
  \centering
  \caption{\textbf{Removing the auxiliary halting loss improves token fitting during adaptation.} Entries report mean logged token losses over matched completed runs: five for memory-only adaptation and three for per-puzzle fine-tuning per setting. Token losses exclude the auxiliary halting loss.}
  \label{tab:cose-cl-token-loss}
  \footnotesize
  \setlength{\tabcolsep}{3pt}
  \smalluncertainties
  \begin{tabular}{@{}llcccc@{}}
    \toprule
    & & \multicolumn{2}{c}{Training stream} & \multicolumn{2}{c}{Public-evaluation stream}\\
    \cmidrule(lr){3-4}\cmidrule(l){5-6}
    Base task memory & Adaptation & $w_h=0.5$ & $w_h=0$ & $w_h=0.5$ & $w_h=0$\\
    \midrule
    \method{} $(r=0)$ & Memory-only & 0.1990 & 0.1650 & 0.2791 & 0.2451\\
    \method{} $(r=0)$ & Per-puzzle FT & 0.0525 & 0.0259 & 0.0597 & 0.0281\\
    \method{} $(r=32)$ & Memory-only & 0.1613 & 0.1266 & 0.2412 & 0.2021\\
    \method{} $(r=32)$ & Per-puzzle FT & 0.0470 & 0.0237 & 0.0552 & 0.0261\\
    \bottomrule
  \end{tabular}
\end{table}

\begin{table}[htbp]
  \centering
  \caption{\textbf{Memory-only adaptation with and without halting loss on ARC-AGI-2 training puzzles.} Scores are means $\pm$ sample standard deviations over completed runs.}
  \label{tab:cose-cl-halt-train-frozen}
  \footnotesize
  \setlength{\tabcolsep}{2.5pt}
  \smalluncertainties
  \begin{tabular}{@{}lcccc@{}}
    \toprule
    & \multicolumn{2}{c}{With halting loss ($w_h=0.5$)} & \multicolumn{2}{c}{Without halting loss ($w_h=0$)}\\
    \cmidrule(lr){2-3}\cmidrule(l){4-5}
    Base task memory & pass@1 & pass@2 & pass@1 & pass@2\\
    \midrule
    \method{} $(r=0,t=32)$ & $17.94\pm1.08$ & $20.60\pm0.86$ & $16.78\pm0.44$ & $20.30\pm0.44$\\
    \method{} $(r=0)$ & $34.76\pm1.10$ & $38.24\pm0.63$ & $36.05\pm1.15$ & $39.73\pm1.48$\\
    \method{} $(r=32,t=32)$ & $22.07\pm1.22$ & $27.42\pm0.48$ & $19.40\pm0.89$ & $23.76\pm0.26$\\
    \method{} $(r=32)$ & $35.26\pm1.44$ & $41.03\pm1.92$ & $38.56\pm0.70$ & $43.58\pm0.58$\\
    \method{} $(r=512)$ & $37.28\pm1.93$ & $42.70\pm0.91$ & $37.81\pm0.77$ & $42.26\pm1.31$\\
    Table $(r=32)$ & $4.33\pm0.24$ & $6.65\pm0.43$ & $5.19\pm0.24$ & $7.77\pm0.24$\\
    Table $(r=512)$ & $14.25\pm0.78$ & $18.47\pm0.39$ & $15.41\pm0.78$ & $19.73\pm0.49$\\
    \bottomrule
  \end{tabular}
\end{table}

\begin{table}[htbp]
  \centering
  \caption{\textbf{Per-puzzle FT Oracle with and without halting loss on ARC-AGI-2 training puzzles.} Scores are means $\pm$ sample standard deviations over completed runs.}
  \label{tab:cose-cl-halt-train-reset}
  \footnotesize
  \setlength{\tabcolsep}{2.5pt}
  \smalluncertainties
  \begin{tabular}{@{}lcccc@{}}
    \toprule
    & \multicolumn{2}{c}{With halting loss ($w_h=0.5$)} & \multicolumn{2}{c}{Without halting loss ($w_h=0$)}\\
    \cmidrule(lr){2-3}\cmidrule(l){4-5}
    Base task memory & pass@1 & pass@2 & pass@1 & pass@2\\
    \midrule
    \method{} $(r=0,t=32)$ & $37.09\pm2.03$ & $40.78\pm2.70$ & $42.93\pm1.22$ & $48.70\pm1.85$\\
    \method{} $(r=0)$ & $34.80\pm0.95$ & $39.18\pm2.02$ & $42.23\pm1.28$ & $47.02\pm2.15$\\
    \method{} $(r=32,t=32)$ & $40.50\pm1.18$ & $45.65\pm0.30$ & $45.23\pm0.83$ & $49.25\pm1.20$\\
    \method{} $(r=32)$ & $36.21\pm1.91$ & $42.54\pm2.15$ & $41.93\pm0.44$ & $46.27\pm1.32$\\
    \method{} $(r=512)$ & $37.71\pm0.53$ & $41.36\pm0.53$ & $43.35\pm1.59$ & $48.51\pm2.02$\\
    Table $(r=32)$ & $33.94\pm3.01$ & $39.38\pm1.84$ & $40.99\pm3.33$ & $46.49\pm0.45$\\
    Table $(r=512)$ & $38.63\pm0.66$ & $43.65\pm1.93$ & $43.40\pm1.02$ & $49.42\pm1.80$\\
    \bottomrule
  \end{tabular}
\end{table}

\begin{table}[htbp]
  \centering
  \caption{\textbf{Joint fine-tuning with and without halting loss on ARC-AGI-2 training puzzles.} Scores are means $\pm$ sample standard deviations.}
  \label{tab:cose-cl-halt-train-joint}
  \footnotesize
  \setlength{\tabcolsep}{2.5pt}
  \smalluncertainties
  \begin{tabular}{@{}lcccc@{}}
    \toprule
    & \multicolumn{2}{c}{With halting loss ($w_h=0.5$)} & \multicolumn{2}{c}{Without halting loss ($w_h=0$)}\\
    \cmidrule(lr){2-3}\cmidrule(l){4-5}
    Base task memory & pass@1 & pass@2 & pass@1 & pass@2\\
    \midrule
    \method{} $(r=0,t=32)$ & $30.58\pm1.61$ & $36.78\pm0.70$ & $26.02\pm2.00$ & $31.88\pm0.58$\\
    \method{} $(r=0)$ & $30.44\pm2.07$ & $35.32\pm1.85$ & $26.31\pm1.85$ & $31.35\pm2.60$\\
    \method{} $(r=32,t=32)$ & $33.95\pm1.20$ & $39.94\pm0.34$ & $29.08\pm0.23$ & $32.83\pm0.45$\\
    \method{} $(r=32)$ & $30.32\pm2.00$ & $36.85\pm2.57$ & $28.06\pm3.98$ & $33.76\pm2.86$\\
    \method{} $(r=512)$ & $33.87\pm2.07$ & $39.77\pm0.71$ & $25.83\pm3.43$ & $31.59\pm4.89$\\
    Table $(r=32)$ & $25.04\pm2.10$ & $31.52\pm2.76$ & $19.10\pm1.50$ & $23.62\pm0.59$\\
    Table $(r=512)$ & $30.64\pm1.33$ & $33.83\pm1.49$ & $18.41\pm3.05$ & $21.49\pm2.79$\\
    \bottomrule
  \end{tabular}
\end{table}

\FloatBarrier

\begin{table}[htbp]
  \centering
  \caption{\textbf{Memory-only adaptation with and without halting loss on ARC-AGI-2 public-evaluation puzzles.} Scores are means $\pm$ sample standard deviations over completed runs.}
  \label{tab:cose-cl-halt-eval-frozen}
  \footnotesize
  \setlength{\tabcolsep}{2.5pt}
  \smalluncertainties
  \begin{tabular}{@{}lcccc@{}}
    \toprule
    & \multicolumn{2}{c}{With halting loss ($w_h=0.5$)} & \multicolumn{2}{c}{Without halting loss ($w_h=0$)}\\
    \cmidrule(lr){2-3}\cmidrule(l){4-5}
    Base task memory & pass@1 & pass@2 & pass@1 & pass@2\\
    \midrule
    \method{} $(r=0,t=32)$ & $1.75\pm0.54$ & $2.54\pm0.57$ & $3.86\pm0.84$ & $5.18\pm0.78$\\
    \method{} $(r=0)$ & $9.30\pm0.48$ & $9.65\pm0.54$ & $9.21\pm1.12$ & $10.76\pm0.96$\\
    \method{} $(r=32,t=32)$ & $5.88\pm0.39$ & $6.58\pm0.31$ & $5.88\pm0.59$ & $6.14\pm0.54$\\
    \method{} $(r=32)$ & $8.74\pm1.97$ & $10.03\pm2.57$ & $11.05\pm1.56$ & $12.19\pm0.90$\\
    \method{} $(r=512)$ & $9.71\pm1.02$ & $11.20\pm0.90$ & $11.46\pm1.08$ & $11.96\pm1.52$\\
    Table $(r=32)$ & $0.00\pm0.00$ & $0.18\pm0.24$ & $0.00\pm0.00$ & $0.09\pm0.20$\\
    Table $(r=512)$ & $3.16\pm0.57$ & $3.60\pm0.65$ & $3.60\pm0.57$ & $4.47\pm0.57$\\
    \bottomrule
  \end{tabular}
\end{table}

\begin{table}[htbp]
  \centering
  \caption{\textbf{Per-puzzle FT Oracle with and without halting loss on ARC-AGI-2 public-evaluation puzzles.} Scores are means $\pm$ sample standard deviations over completed runs.}
  \label{tab:cose-cl-halt-eval-reset}
  \footnotesize
  \setlength{\tabcolsep}{2.5pt}
  \smalluncertainties
  \begin{tabular}{@{}lcccc@{}}
    \toprule
    & \multicolumn{2}{c}{With halting loss ($w_h=0.5$)} & \multicolumn{2}{c}{Without halting loss ($w_h=0$)}\\
    \cmidrule(lr){2-3}\cmidrule(l){4-5}
    Base task memory & pass@1 & pass@2 & pass@1 & pass@2\\
    \midrule
    \method{} $(r=0,t=32)$ & $11.01\pm0.66$ & $12.33\pm1.71$ & $10.58\pm1.18$ & $13.74\pm0.91$\\
    \method{} $(r=0)$ & $8.77\pm1.16$ & $10.77\pm1.69$ & $12.33\pm0.61$ & $14.04\pm1.14$\\
    \method{} $(r=32,t=32)$ & $8.92\pm0.25$ & $11.35\pm0.45$ & $13.74\pm1.01$ & $16.47\pm1.69$\\
    \method{} $(r=32)$ & $10.72\pm1.06$ & $13.01\pm2.44$ & $13.40\pm0.97$ & $15.20\pm0.88$\\
    \method{} $(r=512)$ & $8.70\pm2.38$ & $9.36\pm2.07$ & $13.16\pm0.25$ & $15.06\pm0.51$\\
    Table $(r=32)$ & $8.72\pm0.88$ & $10.48\pm1.56$ & $10.77\pm2.13$ & $13.74\pm1.98$\\
    Table $(r=512)$ & $11.55\pm0.67$ & $12.96\pm0.83$ & $13.89\pm0.76$ & $15.94\pm1.34$\\
    \bottomrule
  \end{tabular}
\end{table}

\begin{table}[htbp]
  \centering
  \caption{\textbf{Joint fine-tuning with and without halting loss on ARC-AGI-2 public-evaluation puzzles.} Scores are means $\pm$ sample standard deviations.}
  \label{tab:cose-cl-halt-eval-joint}
  \footnotesize
  \setlength{\tabcolsep}{2.5pt}
  \smalluncertainties
  \begin{tabular}{@{}lcccc@{}}
    \toprule
    & \multicolumn{2}{c}{With halting loss ($w_h=0.5$)} & \multicolumn{2}{c}{Without halting loss ($w_h=0$)}\\
    \cmidrule(lr){2-3}\cmidrule(l){4-5}
    Base task memory & pass@1 & pass@2 & pass@1 & pass@2\\
    \midrule
    \method{} $(r=0,t=32)$ & $5.95\pm2.47$ & $8.38\pm3.22$ & $6.09\pm1.14$ & $6.77\pm1.85$\\
    \method{} $(r=0)$ & $5.26\pm1.24$ & $6.14\pm0.00$ & $6.07\pm0.10$ & $7.60\pm0.83$\\
    \method{} $(r=32,t=32)$ & $7.36\pm1.24$ & $11.16\pm0.59$ & $5.95\pm1.04$ & $8.63\pm0.15$\\
    \method{} $(r=32)$ & $6.19\pm0.55$ & $8.24\pm1.19$ & $5.75\pm0.94$ & $8.63\pm1.16$\\
    \method{} $(r=512)$ & $6.87\pm1.24$ & $9.72\pm1.55$ & $5.31\pm1.49$ & $7.80\pm1.39$\\
    Table $(r=32)$ & $6.04\pm2.24$ & $7.50\pm2.41$ & $4.24\pm0.25$ & $4.82\pm0.44$\\
    Table $(r=512)$ & $8.43\pm2.60$ & $9.41\pm1.63$ & $4.63\pm0.55$ & $5.17\pm1.48$\\
    \bottomrule
  \end{tabular}
\end{table}

\FloatBarrier

\begin{table}[htbp]
  \centering
  \caption{\textbf{Halting-loss effects on sequential adaptation of the composition module.} All runs use the \method{} $(r=32)$ checkpoint and ARC-AGI-2 training stream, updating shared composition parameters and the current puzzle's private rows. Scores are means $\pm$ sample standard deviations.}
  \label{tab:cose-cl-halt-comp}
  \footnotesize
  \setlength{\tabcolsep}{2.5pt}
  \smalluncertainties
  \begin{tabular}{@{}lccccc@{}}
    \toprule
    Method & $w_h$ & Immediate p@2 & Final p@1 & Final p@2 & ARC-1 after\\
    \midrule
    L2-SP & 0.5 & $41.58\pm1.51$ & $35.48\pm1.80$ & $40.68\pm2.70$ & $79.44\pm0.96$\\
    L2-SP & 0 & $45.40\pm1.68$ & $38.87\pm0.86$ & $44.13\pm0.70$ & $78.33\pm0.00$\\
    Rehearsal & 0.5 & $39.94\pm1.77$ & $31.43\pm0.75$ & $35.50\pm1.04$ & $78.33\pm1.67$\\
    Rehearsal & 0 & $41.94\pm1.25$ & $31.76\pm1.67$ & $37.20\pm1.25$ & $78.06\pm3.76$\\
    EWC & 0.5 & $40.13\pm1.31$ & $14.45\pm6.15$ & $18.10\pm6.58$ & $6.11\pm7.88$\\
    EWC & 0 & $41.99\pm1.99$ & $21.22\pm8.31$ & $25.94\pm8.46$ & $30.28\pm37.33$\\
    Naive & 0.5 & $39.41\pm1.51$ & $8.44\pm2.79$ & $10.59\pm2.83$ & $0.00\pm0.00$\\
    Naive & 0 & $40.72\pm2.12$ & $4.79\pm1.43$ & $6.22\pm2.07$ & $0.00\pm0.00$\\
    \bottomrule
  \end{tabular}
\end{table}

\begin{table}[htbp]
  \centering
  \caption{\textbf{Halting-loss effects on sequential full-model adaptation.} All runs use the \method{} $(r=32)$ checkpoint and ARC-AGI-2 training stream, updating all shared weights and the current puzzle's private rows. Scores are means $\pm$ sample standard deviations; single-run results omit the standard deviation.}
  \label{tab:cose-cl-halt-full}
  \footnotesize
  \setlength{\tabcolsep}{2.5pt}
  \smalluncertainties
  \begin{tabular}{@{}lccccc@{}}
    \toprule
    Method & $w_h$ & Immediate p@2 & Final p@1 & Final p@2 & ARC-1 after\\
    \midrule
    L2-SP & 0.5 & $48.20\pm1.86$ & $1.50\pm2.24$ & $2.07\pm3.23$ & $30.00\pm39.05$\\
    L2-SP & 0 & $52.37\pm2.38$ & $16.50\pm1.33$ & $20.86\pm2.34$ & $77.78\pm0.96$\\
    Rehearsal & 0.5 & $47.90\pm0.26$ & $1.65\pm0.87$ & $1.93\pm1.19$ & $51.67\pm3.33$\\
    Rehearsal & 0 & $45.51\pm0.98$ & $4.58\pm1.18$ & $5.58\pm0.94$ & $56.94\pm4.28$\\
    EWC & 0.5 & $38.40\pm2.02$ & $0.14\pm0.25$ & $0.14\pm0.25$ & $0.00\pm0.00$\\
    EWC & 0 & $44.11\pm0.62$ & $0.79\pm1.36$ & $0.93\pm1.26$ & $6.67\pm11.55$\\
    Naive & 0.5 & --- & $0.00\pm0.00$ & $0.00\pm0.00$ & ---\\
    Naive & 0 & $5.22$ & $0.00\pm0.00$ & $0.00\pm0.00$ & $0.00$\\
    \bottomrule
  \end{tabular}
\end{table}

\FloatBarrier

\section{Synthetic data details}
\label{app:synthetic-data}

This appendix documents the synthetic data used by \sysname{} (Section~\ref{sec:synthetic-data}): the agentic protocol behind the Re-ARC2 dataset, and the ablations that choose the data-mix ratios.

\subsection{Synthetic Re-ARC2 protocol}
\label{app:rearc2}

\paragraph{Per-task generation.}
The ARC-AGI-2 training set contains 1{,}000 puzzles. 391 of them also already appear in the ARC-AGI-1 training set and are therefore covered by the original Re-ARC dataset. For each of the remaining 609 ARC-AGI-2 training tasks, a coding agent (Codex with GPT-5.4 xhigh) is provided with the official task JSON, the project’s Re-ARC DSL reference, and a workflow specification describing the generation process. The agent then produces task-specific generator and verifier programs. The generator directly constructs input-output pairs, while the verifier checks whether candidate pairs satisfy the underlying task rule. Algorithm~\ref{alg:rearc2-workflow} summarizes the workflow.

\paragraph{Acceptance criteria.}
A task package is accepted only if: (i) the verifier exactly reproduces every official pair; (ii) the shared builder produces 1{,}000 unique, non-identity generated examples that all pass verification, unless a finite-support cap is explicitly specified; and (iii) previews of generated examples visually align with the original task family under manual inspection.

\paragraph{Initial human review and correction.}
We generated Re-ARC2 using Codex with GPT-5.4 (xhigh). Human review revealed 19 out of the 609 ARC-AGI-2 training tasks required manual correction. We also tested Claude Code using Opus 4.6 (xhigh), but did not proceed with it because we observed a substantially higher failure rate. We corrected these 19 tasks with the coding agent, 3 additional tasks were repaired because of errors in the official data. Most changes refined palette or geometric support, noise or component statistics, overlaps, or diversity. Only two of them are rule-level failures (\texttt{4e45f183} and \texttt{423a55dc}).

\paragraph{Cross-model audit and final verification.}
After the initial review, we conducted a second round of audit of agent-generated verifiers and generators using a different model family, Claude Opus 5 (xhigh). For each puzzle, the auditing model independently inferred the intended rule from the official examples, checked whether the verifier and generator implemented it, and screened for hard-coded special cases, distributional mismatches, and data errors.

The audit found no issues in 96.4\% of tasks, flagging 22/609 (3.61\%) tasks. As a reference, the same procedure applied to the widely adopted Re-ARC dataset flagged 27/400 (6.75\%) tasks. We manually reviewed all 22 flags and independently assessed them with GPT-5.6 Sol (xhigh). The resulting consensus classifies 9 flags (40.9\%) as errors in the official ARC example pairs (including off-by-one slips, complex outputs with subtle errors), 6 (27.3\%) as false positives involving immaterial edge cases, and 7 (31.8\%) as genuine synthetic-data issues. We corrected all 7 genuine issues and the 9 cases affected by official-example errors. Every correction was verified through human inspection and a third round of fresh Opus 5 audit. Table~\ref{tab:rearc2-corrections} lists the edited puzzle IDs from both rounds.

To ensure a fair comparison, we did not use the corrected official examples when training \sysname{}. These corrections are included only in the Re-ARC2 dataset.

\begin{table}[ht]
  \centering
  \caption{Corrections identified during the two review rounds, grouped by error source.}
  \label{tab:rearc2-corrections}
  \small
  \begin{tabular}{@{}p{0.13\linewidth}p{0.16\linewidth}cp{0.55\linewidth}@{}}
    \toprule
    \raggedright Review round & \raggedright Error source & Count & \raggedright Puzzle IDs \tabularnewline
    \midrule
    \raggedright Round 1 & \raggedright Generator & 19 &
    \raggedright \texttt{15113be4}, \texttt{17cae0c1}, \texttt{2c0b0aff}, \texttt{342ae2ed}, \texttt{3ad05f52}, \texttt{423a55dc}, \texttt{4e45f183}, \texttt{4ff4c9da}, \texttt{538b439f}, \texttt{705a3229}, \texttt{80214e03}, \texttt{8719f442}, \texttt{902510d5}, \texttt{a416fc5b}, \texttt{b7955b3c}, \texttt{c6141b15}, \texttt{c920a713}, \texttt{e6de6e8f}, \texttt{e9fc42f2} \tabularnewline
    \raggedright Round 1 & \raggedright Official data & 3 &
    \raggedright \texttt{b74ca5d1}, \texttt{d931c21c}, \texttt{f18ec8cc} \tabularnewline
    \raggedright Round 2 & \raggedright Generator & 7 &
    \raggedright \texttt{1b8318e3}, \texttt{689c358e}, \texttt{79369cc6}, \texttt{aee291af}, \texttt{c61be7dc}, \texttt{e048c9ed}, \texttt{e5c44e8f} \tabularnewline
    \raggedright Round 2 & \raggedright Official data & 9 &
    \raggedright \texttt{17829a00}, \texttt{85fa5666}, \texttt{963c33f8}, \texttt{a3f84088}, \texttt{ac0c5833}, \texttt{b942fd60}, \texttt{ba1aa698}, \texttt{d753a70b}, \texttt{e1d2900e} \tabularnewline
    \bottomrule
  \end{tabular}
\end{table}

\begin{algorithm}[ht]
  \caption{Agentic workflow for generating one Re-ARC2 task package.}
  \label{alg:rearc2-workflow}
  \begin{algorithmic}[1]
    \Require official task $T$; Re-ARC DSL reference and original Re-ARC implementations
    \Ensure task-specific verifier $V$ and generator $G$; verified synthetic set $D$
    \State Study the official examples of $T$ alongside the DSL reference; infer the task rule and characterize the input distribution
    \State If the rule cannot be inferred reliably, halt and flag $T$ for manual construction
    \State Implement $V$ in Re-ARC verifier style, as a DSL program mapping each input grid to its output grid
    \State Implement $G$ in Re-ARC generator style, sampling the latent structure of an input explicitly and deriving the paired output from it
    \Repeat
      \State Validate $V$ on $T$: every official pair $(x,y)$ must satisfy $V(x)=y$ exactly
      \State Build $D$ with the shared builder: 1{,}000 pairs (or the declared finite-support cap), retaining only unique, non-identity pairs with $V(x)=y$
      \State Re-check every pair in $D$ in an independent verification pass
      \State Render preview sheets of $D$ and compare them visually against the official task family
      \State On any failure, revise $V$ and/or $G$ according to the failed check
    \Until{the official, build, verification, and visual checks all pass}
    \State \Return $V$, $G$, $D$
  \end{algorithmic}
\end{algorithm}

\FloatBarrier

\subsection{Data mix and Re-ARC ratio}
\label{app:data-ratio}

Table~\ref{tab:data-ratio} reports the ablations behind the default \sysname{} data mix (100 Re-ARC augmentations per puzzle with evaluation-task training pairs repeated twice, Appendix~\ref{app:impl}).

\begin{table}[ht]
  \centering
  \caption{\textbf{ARC-AGI-1 runs over Re-ARC augmentation scale and evaluation-example repetition.} The first column is the number of Re-ARC augmentations per Re-ARC puzzle. More augmentations generally help, but it makes evaluation pairs a tiny fraction of each epoch. Repeating evaluation examples partially corrects this imbalance, but too much repetition reduces accuracy.}
  \label{tab:data-ratio}
  \begin{tabular}{cccc}
    \toprule
    Re-ARC aug per puzzle & Evaluation repeats & pass@2 & pass@1000\\
    \midrule
    10  & 1   & 76.38 & 86.25\\
    10  & 2   & 77.38 & 86.62\\
    10  & 5   & 62.12 & 79.25\\
    10  & 10  & 54.62 & 70.50\\
    100 & 1   & 79.25 & 88.62\\
    \textbf{100} & \textbf{2} & \textbf{79.50} & \textbf{89.50}\\
    100 & 5   & 74.75 & 86.75\\
    100 & 10  & 69.25 & 84.88\\
    100 & 20  & 69.12 & 81.62\\
    100 & 50  & 59.38 & 77.75\\
    100 & 100 & 55.38 & 72.38\\
    \bottomrule
  \end{tabular}
\end{table}

\FloatBarrier

\section{Recurrent backbone scaling details}
\label{app:scaling-table}

This appendix provides the visualization and complete results for the two recurrent-backbone scaling studies summarized in Section~\ref{sec:backbone-scaling}.

\begin{figure}[htbp]
  \centering
  \begin{minipage}[t]{0.80\linewidth}
    \centering
    \includegraphics[width=\linewidth]{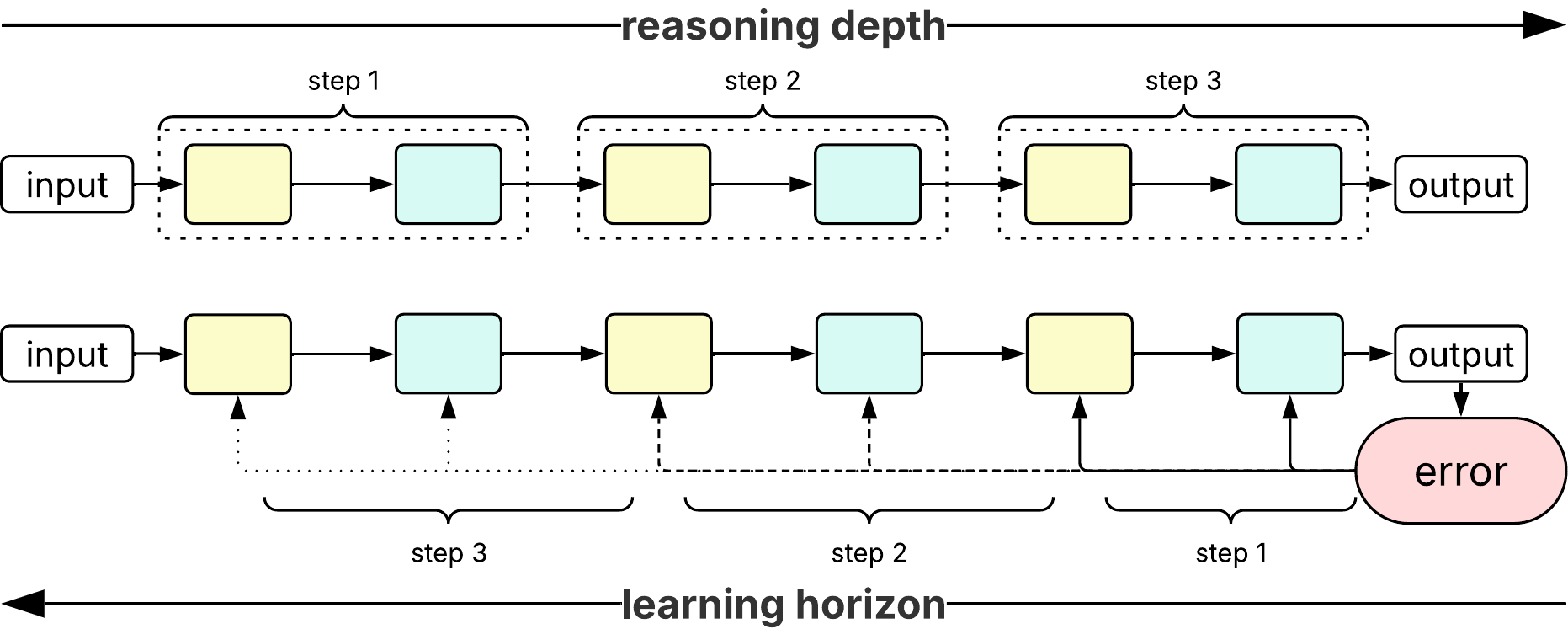}\\
    \footnotesize (a) a diagram for reasoning depth and learning horizon
  \end{minipage}
  \vspace{4pt}

  \begin{minipage}[t]{0.5\linewidth}
    \centering
    \includegraphics[width=\linewidth]{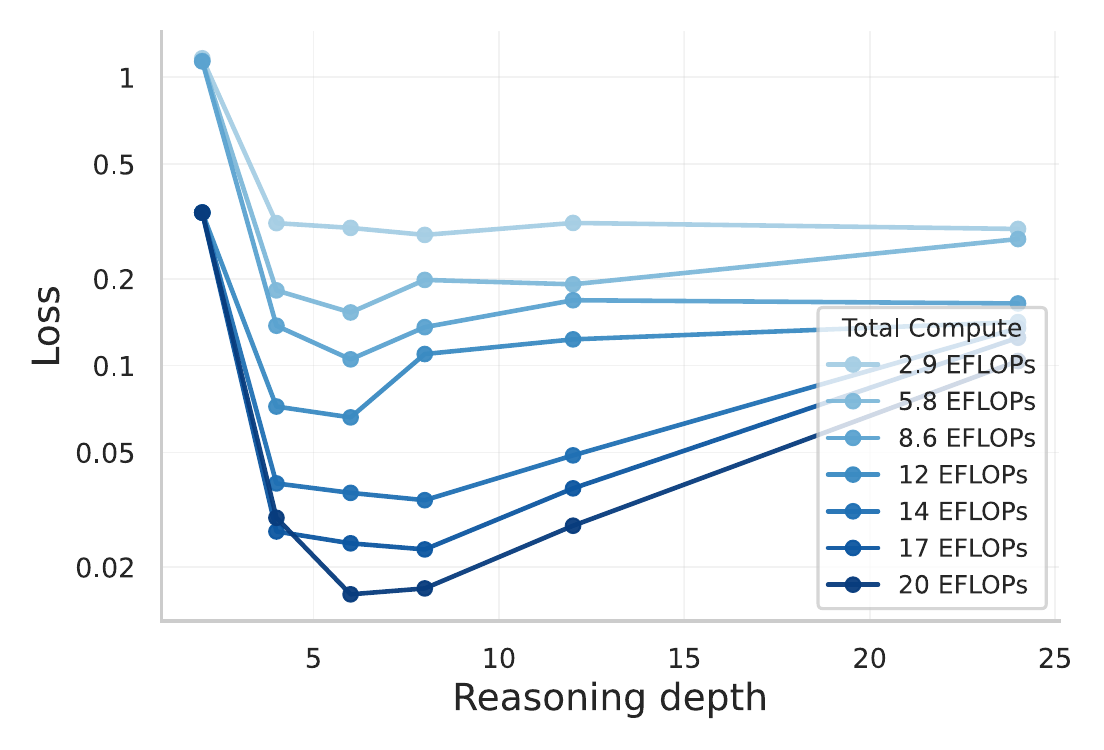}\\
    \footnotesize (b) same block counts with different weight-sharing
  \end{minipage}\hfill
  \begin{minipage}[t]{0.5\linewidth}
    \centering
    \includegraphics[width=\linewidth]{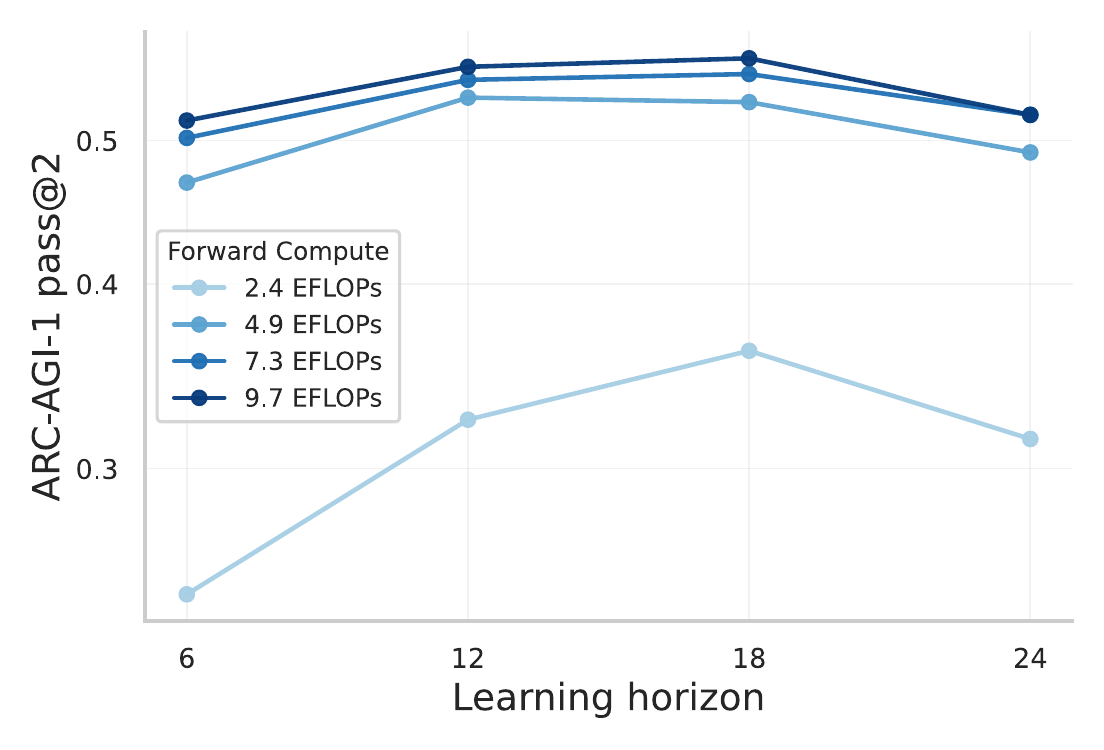}\\
    \footnotesize (c) same network with different learning-horizon
  \end{minipage}
  \caption{\textbf{Balancing reasoning depth and learning horizon in the recurrent backbone.}
  (a) The reasoning blocks can be repeated to produce more iterative refinements, but gradients only supervise the portion covered by the truncated learning horizon.
  (b) At matched FLOPs, using more unique blocks is not automatically better: the strongest settings balance unique block capacity with enough recurrent iterations.
  (c) For a fixed $U{=}2,R{=}12$ backbone, increasing the learning horizon provides more learning signals, notably improving accuracies before gains flatten.
  Studying those tradeoffs has motivated the default recurrent design used in the main system.}
  \label{fig:recurrent-backbone-design}
\end{figure}

The reasoning backbone in XRM consists of a stack of Transformer blocks with weight sharing in the style of Universal Transformer~\citep{dehghani2019universal,gao2025urm}. Specifically, the model contains $U$ unique layers that form a reasoning block, which is repeated $R$ times. We refer to $R$ as the \emph{reasoning depth}. The total number of layers is therefore $L=UR$.
Parameter sharing also enables the use of truncated backpropagation through time (TBPTT). To provide sufficient learning signal, gradients must propagate through at least one full reasoning block. We define the TBPTT horizon $H$ as the \emph{learning horizon}, where $1\leq H \leq R$. When truncation aligns with reasoning-block boundaries, we equivalently report the layer-level learning horizon $H_\ell=UH$. An intuitive illustration is provided in Figure~\ref{fig:recurrent-backbone-design}(a). In a standard transformer, $L=U$, $R=1$ and $H=1$. In contrast, the recurrent structure introduces additional flexibility, allowing us to study how recursive depth and learning horizon affect ARC-AGI performance.

We conducted two scaling studies over recurrent backbone configurations on the ARC-AGI-1 training tasks. The first sweep studies the effect of reasoning depth. Within each run family, we keep the total layer count $L$ fixed and set the TBPTT horizon to half of the network depth ($H_\ell=\frac{1}{2}L$). We vary only the number of unique layers $U$ and the reasoning depth $R$, while maintaining a constant product $L=UR$. This ensures that both forward and backward FLOPs remain identical across all configurations. We log both language modeling loss and evaluation accuracy throughout training.

Figure~\ref{fig:recurrent-backbone-design}(b) illustrates the tradeoff between unique parameters and recursive reasoning in a 48-layer setup, revealing a U-shaped performance curve. Models with too many unique parameters lead to overfitting and memorization for a small dataset such as ARC-AGI. Conversely, models with excessive recursion but too few unique parameters also underperform due to insufficient capacity. These results suggest that strong performance requires balancing iterative refinement depth with an appropriate amount of unique parameter capacity.

The second sweep focuses on the learning horizon. In this experiment, we keep the forward configuration identical, including the total layer count $L$, reasoning depth $R$, and unique layer count $U$, and vary only the TBPTT horizon $H$. While the forward compute remains constant, the backward compute increases proportionally with the learning horizon. Figure~\ref{fig:recurrent-backbone-design}(c) shows the evaluation accuracy for a configuration with $U=2$ and $R=12$. The results indicate that recurrent models require a nontrivial credit-assignment horizon. At one extreme, HRM uses $H=1$~\citep{wang2025hrm}, which provides insufficient learning signal per refinement step and therefore leads to slower learning. However, full backpropagation through all refinement steps is also suboptimal when $R$ is large, as it can degrade generalization and introduce additional training noise along long refinement trajectories.

\FloatBarrier

Within each experiment family, we report the latest evaluation step shared by all runs. Loss is averaged over the 100 logged updates from steps $s-50$ through $s+49$, where $s$ is the reported step.

\begin{table}[ht]
  \centering
  \caption{\textbf{Iso-FLOP recurrent backbone runs on ARC-AGI-1 training tasks.} Within each iso-FLOP family, the total forward depth $L=UR$ and layer-level TBPTT horizon $H_\ell$ are fixed, while the number of unique parameter layers $U$ and recurrent applications $R$ vary. This compares less parameter sharing against deeper recurrence at matched forward compute and learning horizon. Pass@$k$ scores are percentages.}
  \label{tab:scaling}
  \small
  \resizebox{\linewidth}{!}{%
  \begin{tabular}{rrrrrrrrrr}
    \toprule
    $L$ & $U$ & $R$ & $H_\ell$ & Params (M) & Step & Loss & pass@1 & pass@2 & pass@1000\\
    \midrule
    12 & 2  & 6  & 6  & 6.8  & 67588  & 0.012 & 47.50 & 52.08 & 71.38\\
    12 & 3  & 4  & 6  & 10.2 & 67588  & 0.009 & 45.88 & 50.21 & 69.75\\
    12 & 4  & 3  & 6  & 13.6 & 67588  & 0.006 & 40.83 & 46.08 & 66.62\\
    12 & 6  & 2  & 6  & 20.5 & 67588  & 0.005 & 34.62 & 39.21 & 54.62\\
    \midrule
    24 & 2  & 12 & 12 & 6.8  & 33794  & 0.034 & 47.00 & 52.96 & 69.75\\
    24 & 3  & 8  & 12 & 10.2 & 33794  & 0.029 & 48.12 & 52.08 & 70.12\\
    24 & 4  & 6  & 12 & 13.6 & 33794  & 0.027 & 46.62 & 50.96 & 66.50\\
    24 & 6  & 4  & 12 & 20.5 & 33794  & 0.024 & 43.00 & 47.21 & 63.88\\
    24 & 8  & 3  & 12 & 27.3 & 33794  & 0.015 & 36.75 & 41.21 & 58.13\\
    24 & 12 & 2  & 12 & 40.9 & 33794  & 0.033 & 31.79 & 36.88 & 51.12\\
    \midrule
    48 & 2  & 24 & 24 & 6.8  & 16897  & 0.175 & 31.79 & 36.46 & 59.25\\
    48 & 4  & 12 & 24 & 13.6 & 16897  & 0.137 & 41.33 & 44.04 & 61.38\\
    48 & 6  & 8  & 24 & 20.5 & 16897  & 0.109 & 40.33 & 46.00 & 60.17\\
    48 & 8  & 6  & 24 & 27.3 & 16897  & 0.093 & 38.71 & 44.25 & 58.92\\
    48 & 12 & 4  & 24 & 40.9 & 16897  & 0.103 & 35.71 & 40.50 & 53.67\\
    48 & 24 & 2  & 24 & 81.8 & 16897  & 0.573 & 0.00  & 0.00  & 0.62\\
    \bottomrule
  \end{tabular}%
  }
\end{table}

\begin{table}[ht]
  \centering
  \caption{\textbf{Learning horizon sweep at fixed forward configuration on ARC-AGI-1 training tasks.} All rows use total depth $L=24$; within each backbone shape, $U$ and $R$ are fixed while the layer-level learning horizon $H_\ell$ varies. This tests how much of the recurrent computation should receive gradient updates during training, separate from the forward compute used at evaluation. Pass@$k$ scores are percentages.}
  \label{tab:scaling-tbptt}
  \small
  \resizebox{\linewidth}{!}{%
  \begin{tabular}{rrrrrrrrrr}
    \toprule
    $U$ & $R$ & $L$ & $H_\ell$ & Params (M) & Step & Loss & pass@1 & pass@2 & pass@1000\\
    \midrule
    2  & 12 & 24 & 6  & 6.8  & 67588  & 0.014 & 46.25 & 51.58 & 70.62\\
    2  & 12 & 24 & 12 & 6.8  & 67588  & 0.009 & 49.75 & 56.08 & 71.38\\
    2  & 12 & 24 & 18 & 6.8  & 67588  & 0.007 & 51.88 & 56.83 & 71.12\\
    2  & 12 & 24 & 24 & 6.8  & 67588  & 0.017 & 47.00 & 52.04 & 69.75\\
    \midrule
    3  & 8  & 24 & 3  & 10.2 & 67588  & 0.015 & 39.25 & 45.83 & 66.75\\
    3  & 8  & 24 & 6  & 10.2 & 67588  & 0.012 & 40.88 & 47.83 & 68.25\\
    3  & 8  & 24 & 12 & 10.2 & 67588  & 0.009 & 50.38 & 52.71 & 70.50\\
    3  & 8  & 24 & 18 & 10.2 & 67588  & 0.009 & 48.12 & 53.83 & 69.62\\
    3  & 8  & 24 & 24 & 10.2 & 67588  & 0.007 & 42.50 & 49.33 & 65.62\\
    \midrule
    4  & 6  & 24 & 4  & 13.6 & 67588  & 0.011 & 37.92 & 42.50 & 60.00\\
    4  & 6  & 24 & 8  & 13.6 & 67588  & 0.011 & 43.88 & 48.33 & 68.75\\
    4  & 6  & 24 & 12 & 13.6 & 67588  & 0.008 & 49.00 & 53.08 & 70.62\\
    \midrule
    6  & 4  & 24 & 6  & 20.5 & 50691  & 0.009 & 40.00 & 44.71 & 62.12\\
    6  & 4  & 24 & 12 & 20.5 & 50691  & 0.012 & 45.88 & 50.83 & 67.25\\
    6  & 4  & 24 & 18 & 20.5 & 50691  & 0.009 & 46.88 & 52.08 & 68.00\\
    6  & 4  & 24 & 24 & 20.5 & 50691  & 0.008 & 39.83 & 45.29 & 62.88\\
    \midrule
    12 & 2  & 24 & 12 & 40.9 & 33794  & 0.033 & 31.50 & 37.46 & 50.38\\
    12 & 2  & 24 & 16 & 40.9 & 33794  & 0.019 & 35.88 & 41.96 & 54.87\\
    12 & 2  & 24 & 20 & 40.9 & 33794  & 0.018 & 36.62 & 42.21 & 59.79\\
    12 & 2  & 24 & 24 & 40.9 & 33794  & 0.021 & 36.00 & 41.33 & 56.25\\
    \bottomrule
  \end{tabular}%
  }
\end{table}

\FloatBarrier

\section{Inference and evaluation protocols}
\label{app:ttc}

This appendix details the evaluator behind the inference-time aggregation results in Section~\ref{sec:ttc-main}. Table~\ref{tab:ttc-protocols} compares the aggregation choices made by prior XRM evaluators and by the standard, additional-compute, and continual-learning protocols used in this work. Table~\ref{tab:ttc} then isolates the accuracy gain from our additional test-time compute.

\begin{table}[ht]
  \centering
  \caption{\textbf{Evaluation protocols differ in view aggregation, checkpoint aggregation, and refinement depth.} Relative evaluation compute estimates the inference cost from the number of views, evaluated checkpoints, and refinement steps, normalized to No TTC ($=1$). Exponential vote decay uses a half-life of 10 checkpoints. }
  \label{tab:ttc-protocols}
  \small
  {\setlength{\tabcolsep}{3pt}%
  \renewcommand{\arraystretch}{1.15}%
  \begin{tabularx}{\linewidth}{@{}>{\raggedright\arraybackslash}p{0.22\linewidth}*{5}{>{\centering\arraybackslash}X}@{}}
    \toprule
    Protocol & \shortstack{Augmentation\\view voting} & \shortstack{Checkpoint\\vote window} & \shortstack{Exponential\\vote decay} & \shortstack{Outer-loop\\steps} & \shortstack{Relative \\compute}\\
    \midrule
    HRM/TRM~\citep{wang2025hrm,jolicoeur2025trm} & Yes & All & No & 16 & $\sim 1\times10^{4}$\\
    URM~\citep{gao2025urm} & Yes & All & No & 16, 24 & $\sim 1\times10^{5}$\\
    \midrule
    No TTC & No & Current only & No & 16 & $1$\\
    Standard protocol (ours) & Yes & 10 most recent & No & 16 & $\sim 1\times10^{4}$\\
    Additional TTC (ours) & Yes & recent half & Yes & 16, 24 & $\sim 6\times10^{4}$\\
    Continual learning (ours) & Yes & Current only & No & 16 & $\sim 1\times10^{3}$\\
    \bottomrule
  \end{tabularx}%
  }
\end{table}

\begin{table}[ht]
  \centering
  \caption{\textbf{Checkpoint replay and deeper refinement improve pass@2.} The replay evaluator aggregates votes from recent checkpoints with exponential decay and ensembles 16 and 24-step outer refinement loops, compared with the 16-step default evaluator.}
  \label{tab:ttc}
  \begin{tabular}{llccc}
    \toprule
    Benchmark & Evaluator & pass@1 & pass@2 & pass@1000\\
    \midrule
    ARC-AGI-1 & \method{}, Re-ARC (aug=100) & 74.75 & 79.62& 89.25\\
    ARC-AGI-1 & + loops 16/24, replay $h_{1/2}{=}10$ & \textbf{75.75} & \textbf{81.50}& \textbf{91.25}\\
    \midrule
    ARC-AGI-2 & \method{}, synthetic Re-ARC2 (aug=100) & \textbf{34.72} & 38.47& 55.0\\
    ARC-AGI-2 & + loops 16/24, replay $h_{1/2}{=}10$ & 33.50 & \textbf{39.30}& \textbf{58.3}\\
    \bottomrule
  \end{tabular}
\end{table}

The evaluation protocol used in TRM already includes several test-time compute techniques to improve performance~\citep{jolicoeur2025trm}. In the reasoning backbone, early stopping is disabled and the outer refinement loop is fixed at 16 iterations to enable higher-quality predictions. At the ensemble level, predictions from augmented puzzle views are inverted back to the original coordinate system, allowing multiple predictions to be aggregated for the same puzzle. Vote counts and confidence scores are then combined across all augmented views, and pass@k is reported as the fraction of test inputs whose correct output appears within the top-k voted predictions. The voting mechanism also aggregates predictions from all previous checkpoints.

We further introduce an exponential decay over previous checkpoints to reduce the influence of lower-quality predictions from earlier stages of training. Specifically, we apply a decay schedule with a half-life of 10 recent checkpoints. Inspired by the practices used in URM, we additionally ensemble predictions generated using an extended outer refinement loop of 24 iterations~\citep{gao2025urm}. With these additional test-time compute modifications, performance on both ARC-AGI-1 and ARC-AGI-2 improves by approximately 1\% relative to the original evaluation protocol, as shown in Table~\ref{tab:ttc}.

However, we observe that the benefits of additional test-time compute diminish with extended training as the models approach convergence. Consequently, the best results reported in Figure~\ref{fig:overall-arc-performance} for the extended training run are evaluated using the standard test-time compute protocol, without additional iterations or exponential decay.

\FloatBarrier

\section{\sysname{} implementation and optimization details}
\label{app:impl}

\subsection{System configuration}

All ARC-AGI experiments in this paper use the configuration below unless otherwise stated. Most of the design choices follow the URM~\citep{gao2025urm}.

\paragraph{Input encoding and task prefix.}
Each input and output grid is placed on a $30\times30$ canvas and flattened in row-major order, which yields 900 grid tokens. We use a 12-token vocabulary consisting of padding, an end-of-grid marker, and the ten ARC colors. Colors $0,\ldots,9$ are mapped to tokens $2,\ldots,11$; token 0 is padding and token 1 forms an L-shaped boundary immediately below and to the right of the grid. Padding targets are excluded from the prediction loss. During training, a grid is placed at the origin with probability 0.2 and otherwise translated uniformly over positions for which both the grid and its boundary fit on the canvas. Evaluation uses the origin placement.

The 900 grid tokens are preceded by a 16-position task prefix. The 512-dimensional task embedding occupies the first prefix position and the remaining 15 positions are zero. The prediction head discards the prefix and produces one output distribution for each of the 900 canvas positions.

\paragraph{Recurrent backbone.}
The backbone has hidden width $d=512$, 8 attention heads, and $U=4$ unique Transformer layers. Each layer consists of non-causal self-attention followed by a depthwise-convolutional SwiGLU block (introduced in the URM), with residual connections and RMS normalization. The feed-forward expansion factor is 4, giving intermediate width 1536, and the depthwise convolution has kernel size two. We use one-dimensional RoPE with base 10000 for the reported models; two-dimensional RoPE (introduced in VARC~\citep{hu2025varc}) is implemented as an architectural ablation but is not used by the final configurations.

The four-layer block is applied for $R=12$ recurrent steps, yielding 48 Transformer-layer applications per inner forward pass. Input embeddings are reinjected before each recurrent application. Training uses truncated backpropagation through the final $H=6$ recurrent steps. 

\paragraph{Task-memory architecture.}
The reported models use one FiLM-concat compositional head. Puzzle and dihedral embeddings have width 512. The slot-permutation color representation divides this width into a 53-dimensional shared base and nine 51-dimensional color slots. The concatenated dihedral and color representations are passed through separate projections to produce 512-dimensional FiLM scale and shift vectors, which modulate the puzzle embedding as in Equation~\ref{eq:cose-comp}.

The private residual table uses rank $r=32$ and a bias-free linear projection back to width 512. All factor lookup vectors and private residual rows are initialized to zero. ARC-AGI-1 uses the gated residual in Equation~\ref{eq:cose-final}; its gate is initialized with zero weights. ARC-AGI-2 uses the ungated residual. The resulting 512-dimensional vector is inserted into the first task-prefix position described above.

\paragraph{Outer-loop refinement and loss.}
In addition to the 12-step inner recurrence, each sample is processed for up to 16 outer refinement steps. The latent state is detached between outer steps, so each step provides a separate supervised prediction while carrying forward the current representation. The model has a token-prediction head and a scalar halt/confidence head read from the first prefix position.

Let $\mathcal L_{\rm tok}$ be stablemax cross-entropy over the non-padding output tokens. The halt target is 1 when the entire predicted output is correct at the current refinement step. We optimize
\begin{equation}
  \mathcal L
  = \mathcal L_{\rm tok}
  + \tfrac12\mathcal L_{\rm halt},
\end{equation}
where $\mathcal L_{\rm halt}$ is binary cross-entropy. During training, a sample may halt when its halt logit becomes positive. With probability 0.1, we instead impose a uniformly sampled minimum halt step between 2 and 16 to encourage exploration of later refinements. Evaluation disables early halting and performs all 16 steps by default. The halt probability is also used as the confidence tie-breaker when predictions receive the same vote count.

\paragraph{Training data and sampling.}
The training mixture contains the official ARC-AGI training tasks, the training pairs of the public evaluation tasks, ConceptARC, and the synthetic set: Re-ARC for ARC-AGI-1 and Re-ARC2 for ARC-AGI-2. Held-out outputs from public evaluation tasks are never used for optimization.

For official ARC, ConceptARC, and evaluation tasks, we generate up to 1000 distinct offline dihedral--color views per puzzle. Each Re-ARC or Re-ARC2 task contains 1000 generated pairs and up to 100 offline views. The ConceptARC and evaluation-task demonstrations are each repeated twice in the sampling index, while the official and synthetic sources are included once. Samples are shuffled across the combined index; canvas translation is sampled online.

\paragraph{Optimization.}
The effective global batch size is 768. Private residual rows, and the corresponding rows in independent-table baselines, are updated with distributed SignSGD at learning rate $10^{-2}$ and decoupled weight decay 0.1. For the final full-system experiments, two-dimensional dense tensors are optimized with Muon~\citep{jordan2024muon}; remaining dense parameters use auxiliary AdamW. Muon uses momentum 0.95, Nesterov momentum, and five Newton--Schulz iterations. AdamW uses $\beta=(0.9,0.95)$ and $\epsilon=10^{-8}$. Both dense parameter groups use learning rate $10^{-4}$ and weight decay 0.1.

Both learning rates receive a linear warmup over the first 2000 updates and remain constant thereafter. We apply no gradient clipping. Evaluation uses an exponential moving average of the dense parameters with decay 0.999; the sparse task rows retain their current values. In the AdamW rows of Table~\ref{tab:optimizer-ablations}, AdamW replaces Muon for two-dimensional dense tensors, while the sparse-row optimizer remains unchanged.

\paragraph{Training and evaluation budgets.}
We train the standard ARC-AGI-1 and ARC-AGI-2 configurations for 600k and 1M updates, respectively, saving and evaluating an EMA checkpoint every 10k updates. The extended runs continue the same configurations to 1.6M and 2.38M updates. ARC-AGI-1 uses gated rank-32 \method{}, whereas ARC-AGI-2 uses the ungated rank-32 variant. The other test-time compute setups are specified in Appendix~\ref{app:ttc}. Some architectural ablations on ARC-AGI-1 use the training-task split, optimizing its demonstration pairs and evaluating its held-out pairs to reduce compute cost.

\subsection{Optimizer comparison}
\label{app:optimizer-ablations}

Table~\ref{tab:optimizer-ablations} compares AdamW and Muon on ARC-AGI-1 across Re-ARC data mixes. The two optimizers reach comparable saturated accuracy, but Muon often converges faster, which motivates its use for the longest ARC-AGI runs.

\begin{table}[ht]
  \centering
  \caption{\textbf{Optimizer ablation on ARC-AGI-1.} AdamW and Muon reach comparable saturated accuracy across Re-ARC data mixes, while Muon often reaches its best checkpoint in fewer training steps. The 100/2 Muon row reports mean $\pm$ sample standard deviation over three runs; all other rows are single runs.}
  \label{tab:optimizer-ablations}
  \small
  \resizebox{\linewidth}{!}{%
  \smalluncertainties
  \begin{tabular}{llcccc}
    \toprule
    Re-ARC aug count & Eval repeats & Optimizer & pass@1 & pass@2 & pass@1000\\
    \midrule
    10 & 1 & AdamW & 70.88 & 76.38 & 86.25\\
    10 & 1 & Muon  & 71.38 & 78.38 & 90.25\\
    100 & 1 & AdamW & 72.62 & 79.25 & 88.62\\
    100 & 1 & Muon  & 72.50 & 77.38 & 91.50\\
    100 & 2 & AdamW & 74.12 & 79.50 & 89.50\\
    100 & 2 & Muon  & $74.29 \pm 0.75$ & $80.71 \pm 0.51$ & $90.62 \pm 0.50$\\
    \bottomrule
  \end{tabular}%
  }
\end{table}

\FloatBarrier

\section{Compute resource estimates}
\label{app:compute}

This section reports estimated training costs for our \sysname{} model on different hardware platforms. The single-node measurements in Table~\ref{tab:single-node-compute} are intended to provide realistic cost estimates for conventional training hardware (rented from vast.ai). The experiments presented in this paper were conducted on ALCF systems; we also include corresponding cost estimates for those platforms in Table~\ref{tab:alcf-compute-scaling} for reference. We do not report the actual wall-clock training time for the production runs because they were executed across multiple queued job segments with different node counts and configurations.

For a training run, we estimate
\begin{equation}
  T_\text{run}
  = t_\text{train}N_\text{train}
  + t_\text{eval}N_\text{eval-iters}N_\text{evals}.
\end{equation}
The ARC-AGI-1 \sysname{} run uses 600k training steps, evaluation every 10k steps, 60 evaluations total, and 502 evaluation iterations per evaluation. The ARC-AGI-2 \sysname{} run uses 1M training steps, 100 evaluations, and 216 evaluation iterations per evaluation.

\begin{table}[ht]
  \centering
  \caption{\textbf{Single-node compute estimates for ARC training runs.} Train and eval rows report seconds per training step and per evaluation iteration, total rows report projected end-to-end wall-clock hours for the full training. }
  \label{tab:single-node-compute}
  \small
  \begin{tabular}{lllrrr}
    \toprule
    & & & \multicolumn{3}{c}{Hardware (1 node, 8 GPUs)}\\
    \cmidrule(lr){4-6}
    Benchmark & Model & Metric & RTX PRO 6000 & H100 & B200\\
    \midrule
    ARC-AGI-1 & TRM               & Train (s/step) & 0.195 & 0.110 & 0.061\\
              &                   & Eval (s/iter)  & 1.81   & 1.74   & 0.525\\
              &                   & Total (h)      & 47.6   & 32.9   & 14.6\\
    \cmidrule(lr){2-6}
              & URM               & Train (s/step) & 0.578 & 0.385 & 0.257\\
              &                   & Eval (s/iter)  & 4.09   & 2.70   & 1.78\\
              &                   & Total (h)      & 130.5  & 86.8   & 57.7\\
    \cmidrule(lr){2-6}
              & \textbf{\sysname{}} & Train (s/step) & 0.585 & 0.390 & 0.260\\
              &                   & Eval (s/iter)  & 4.09   & 2.70   & 1.78\\
              &                   & Total (h)      & 131.7  & 87.6   & 58.2\\
    \midrule
    ARC-AGI-2 & URM               & Train (s/step) & 0.578 & 0.390 & 0.260\\
              &                   & Eval (s/iter)  & 4.09   & 2.70   & 1.78\\
              &                   & Total (h)      & 185.0  & 124.5  & 82.9\\
    \cmidrule(lr){2-6}
              & \textbf{\sysname{}} & Train (s/step) & 0.585 & 0.400 & 0.260\\
              &                   & Eval (s/iter)  & 4.09   & 2.70   & 1.78\\
              &                   & Total (h)      & 187.0  & 127.3  & 82.9\\
    \bottomrule
  \end{tabular}
\end{table}

\begin{table}[ht]
  \centering
  \caption{\textbf{ALCF compute estimates for ARC training runs.} Rows report measured seconds per training update and evaluation iteration on Polaris and Aurora, followed by projected end-to-end hours for the full training run. Polaris uses 4 NVIDIA A100 GPUs per node, Aurora uses 6 Intel Data Center GPU Max Series GPUs per node. Aurora 24 and 32-node \sysname{} experiments failed deterministically with driver-level segmentation faults, reported directly as ``segfault.''}
  \label{tab:alcf-compute-scaling}
  \scriptsize
  \resizebox{\linewidth}{!}{%
  \begin{tabular}{llrrrrrrrrr}
    \toprule
    & & \multicolumn{3}{c}{ARC-AGI-1 TRM} & \multicolumn{3}{c}{ARC-AGI-1 \textbf{\sysname{}}} & \multicolumn{3}{c}{ARC-AGI-2 \textbf{\sysname{}}}\\
    \cmidrule(lr){3-5}\cmidrule(lr){6-8}\cmidrule(lr){9-11}
    System & Nodes & Train (s) & Eval (s) & Total (h) & Train (s) & Eval (s) & Total (h) & Train (s) & Eval (s) & Total (h)\\
    \midrule
    Polaris & 1  & 0.535 & 5.00 & 131.0 & 1.740 & 12.50 & 394.6 & 1.750 & 12.60 & 561.7\\
            & 4  & 0.188 & 1.26 & 41.9  & 0.550 & 3.15  & 118.0 & 0.550 & 3.15  & 171.7\\
            & 8  & 0.107 & 0.69 & 23.6  & 0.320 & 1.65  & 67.1  & 0.316 & 1.64  & 97.8\\
    \midrule
    Aurora  & 1  & 1.630 & 15.36 & 400.2 & 3.720 & 25.46 & 833.0 & 3.720 & 25.46 & 1186.1\\
            & 4  & 0.413 & 3.79  & 100.5 & 1.040 & 6.28  & 225.9 & 1.040 & 6.28  & 326.6\\
            & 8  & 0.230 & 1.85  & 53.8  & 0.550 & 3.12  & 117.8 & 0.550 & 3.12  & 171.5\\
            & 16 & 0.160 & 1.03  & 35.3  & 0.410 & 1.61  & 81.8  & 0.410 & 1.61  & 123.5\\
            & 20 & 0.150 & 0.73  & 31.1  & 0.360 & 1.21  & 70.1  & 0.360 & 1.21  & 107.3\\
            & 24 & 0.140 & 0.55  & 27.9  & \multicolumn{3}{c}{segfault} & \multicolumn{3}{c}{segfault}\\
            & 32 & 0.150 & 0.56  & 29.7  & \multicolumn{3}{c}{segfault} & \multicolumn{3}{c}{segfault}\\
            & 36 & 0.150 & 0.45  & 28.8  & 0.408 & 0.84  & 75.0  & 0.408 & 0.84  & 118.4\\
            & 48 & 0.170 & 0.44  & 32.0  & 0.510 & 0.83  & 91.9  & 0.510 & 0.83  & 146.6\\
            & 64 & 0.212 & 0.452 & 39.1  & 0.625 & 0.826 & 111.1 & 0.625 & 0.826 & 178.6\\
    \bottomrule
  \end{tabular}%
  }
\end{table}

\end{document}